# A review of laser scanning for geological and geotechnical applications in underground mining

Sarvesh Kumar Singh[1], Bikram Pratap Banerjee[1,2], Simit Raval[1,*]

[1]School of Minerals and Energy Resources Engineering, University of New South Wales, Sydney, NSW 2052, Australia
[2]Agriculture Victoria, Grains Innovation Park, 110 Natimuk Road, Horsham, VIC 3400, Australia
Email: sarveshkumar.singh@student.unsw.edu.au, b.banerjee@unswalumni.com, [*]simit@unsw.edu.au
*Corresponding author

## Abstract

Laser scanning can provide timely assessments of mine sites despite adverse challenges in the operational environment. Although there are several published articles on laser scanning, there is a need to review them in the context of underground mining applications. To this end, a holistic review of laser scanning is presented including progress in 3D scanning systems, data capture/processing techniques and primary applications in underground mines. Laser scanning technology has advanced significantly in terms of mobility and mapping, but there are constraints in coherent and consistent data collection at certain mines due to feature deficiency, dynamics, and environmental influences such as dust and water. Studies suggest that laser scanning has matured over the years for change detection, clearance measurements and structure mapping applications. However, there is scope for improvements in lithology identification, surface parameter measurements, logistic tracking and autonomous navigation. Laser scanning has the potential to provide real-time solutions but the lack of infrastructure in underground mines for data transfer, geodetic networking and processing capacity remain limiting factors. Nevertheless, laser scanners are becoming an integral part of mine automation thanks to their affordability, accuracy and mobility, which should support their widespread usage in years to come.

# 1 Introduction

Advancements and miniaturisation in electro-optical sensor systems have propelled the development of mobile laser scanning systems based on light detection and ranging (LiDAR) principles. These systems have been extensively used in numerous applications that include monitoring deformation, detecting changes, mapping utilities, navigating vehicles and automating industry [1,2]. Industries such as transport, agriculture, energy and manufacturing have greatly benefited from such sensing technologies. Over recent years, the mining industry has increasingly been exploring possibilities to adopt digital sensing technologies to gain comprehensive information from high-resolution and large-scale imaging. However, the applicability of multi-sensor technologies in underground mines has been limited due to (1) a lack of availability of a spatial reference framework, (2) intrinsic-safety related hazards for certain mines (particularly coal mines) and (3) uneven terrain and dusty conditions [3].

Laser sensors are becoming increasingly affordable due to the recent boost in autonomous and driverless vehicles that require 3D spatial information to successfully operate. These sensors provide rich geometric information through rapid scanning, in the form of a 3D point cloud, which can be used to digitally reconstruct the environment. The current means of producing point clouds in underground mines mainly involve terrestrial laser scanning (TLS) and mobile laser scanning (MLS) [3,4]. In the former, the environment is scanned from a fixed point, while in the latter, mobility is achieved using simultaneous localisation and mapping (SLAM) algorithms for scanning. MLS offers several benefits over TLS in terms of simplicity, scale and avoiding blind spots during data capture. Researchers are continually trying to improve the mapping and localisation of MLS; still, the accuracy remains mostly in centimetres. Currently far superior, TLS can provide up to millimetre-level accuracy [1,4]. Nevertheless, the sensor mobility of MLS is an important aspect in terms of efficiency that affords it a range of applications that had previously been challenging in mining when using stationary sensing.

Accuracy requirements dictate the decision made when selecting a laser scanner in underground mines for a particular application. To monitor a small area with high accuracy, such as for convergence measurement, TLS is suitable [5,6]. Meanwhile, applications favour MLS when they require rapid and routine monitoring of a large area, such as for change detection, deformation monitoring, object detection or localisation, for which the permissible error may only be a few centimetres [7,8]. In terms of operational value, mine operators increasingly request MLS solutions as they are easy to operate, quickly scan large areas and facilitate wider applications and scanning options (e.g. human backpack, vehicle- and drone-mounted).

This research review aimed to identify the challenges with acquiring accurate point clouds in underground mines and review the suitability of point clouds, obtained either via TLS, MLS or structure from motion (SfM), for relevant geological and geotechnical mining applications. The rest of the paper is arranged into six sections. Section 2 outlines the method we followed to filter relevant studies and sets out the statistical meta-analysis of published research. Section 3 reviews the factors associated with mapping and mobilities in underground mines to identify challenges and future areas of potential. Section 4 investigates how we can collect coherent and spatially referenced multi-temporal 3D point cloud data in an environment lacking a global navigation satellite system (GNSS) signal. Section 5 provides a critical review of all the major geological and geotechnical applications for laser scanning in underground mines. This section also reviews the processing algorithms for selected applications and provides insights into the scope for future improvements. Section 6 explores relevant potential applications for laser scanning technology where a current issue may be addressed and also proposes future research directions. Finally, concluding remarks are presented in Section 7. The reviewed applications were selected based on their relevance to mining in regards to construction, planning, safety/hazard management and temporal monitoring. The underground environment we analysed concerns GNSS-denied caves, tunnels and underground mines.

## 2 Research methods

Though a considerable number of published articles employed laser scanning, only a few reviewed its relevance in the underground mining context [8,9]. To address the research gap, this review paper describes the current state of laser scanning in holistic mining applications. A systematic review was conducted through an extensive search of various research databases including Web of Science, Scopus, the American Society of Civil Engineers (ASCE) library and IEEE Xplore. The searched contents included peer-reviewed journal articles, conference papers, books, magazines/bulletins and scientific reports, all published between 2000 and 2021, found using relevant keywords. From Web of Science, Scopus, the ASCE library and IEEE Xplore, we found 79, 267, 18 and 13 articles, respectively, with some overlap between them. Our search results were further refined by applying filtering criteria, using constraints such as (1) focusing on the GNSS-denied environment, (2) requiring geotechnical or geological components and (3) concerning applications within the main hazard categories identified by mine safety bodies (Fig. 1), as well as potentially benefiting the mining industry.

| Mining domains relevant for laser scanning | |
|---|---|
| 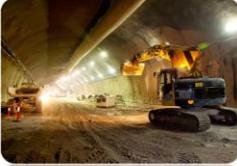 | **Mine ventilation and rescue plan**<br>✓ Direction and quality of ventilation flow<br>✓ Location of ventilation controls<br>✓ Mine ventilation factors such as friction |
| 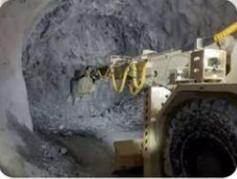 | **Mine excavation**<br>✓ Monitor shaft, stope, pass, trench, costean and pit<br>✓ Analyse structural discontinuities<br>✓ Monitor temporal changes during and after excavation |
| 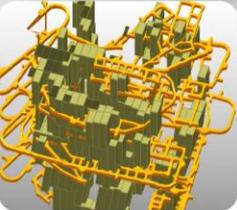 | **Mine plan layout, design and construction**<br>✓ Survey grid system<br>✓ Geological, geotechnical, meteorological and topographical data<br>✓ Access and travel way<br>✓ Stockpile, dumps and tailing dams<br>✓ Vehicle interaction |
| 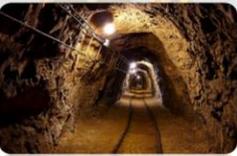 | **Mine roads and rail haulage**<br>✓ Changes in road haulage<br>✓ Temporal monitoring of mine roads |
| 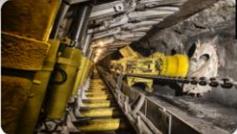 | **Monitor logistics, hazardous substances and dangerous goods**<br>✓ Detect misuse, theft or loss of substances and good<br>✓ Detect leaks, spills and unintended emissions of substances and goods |
| 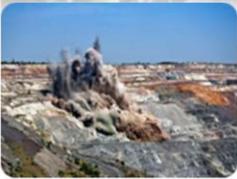 | **Identify interaction hazards before/after using explosives**<br>✓ Analysis of unstable ground<br>✓ Triggering of secondary dust or gas explosion<br>✓ After blast change/ deformation monitoring<br>✓ Fragmentation analysis |

**Fig. 1.** Mining domains and corresponding applications relevant to laser scanning.

A meta-analysis of the reviewed literature identified China, the United States, Canada and Australia as the leading countries actively publishing work on the applied use of laser scanning in underground mining (Fig. 2). Evidently, these countries are also leaders in terms of resource extraction through underground coal and metal mining. The statistics shown in Fig. 2 concern only published case studies that were scientifically peer-reviewed and should not be conflated with unpublished practical use cases of laser scanning in underground mines. A yearly publication plot illustrates how, since 2010, increasing numbers of use cases have been published for laser scanners in underground mines, with a recent drop since 2019, likely due to the COVID-19 pandemic (Fig. 2b).

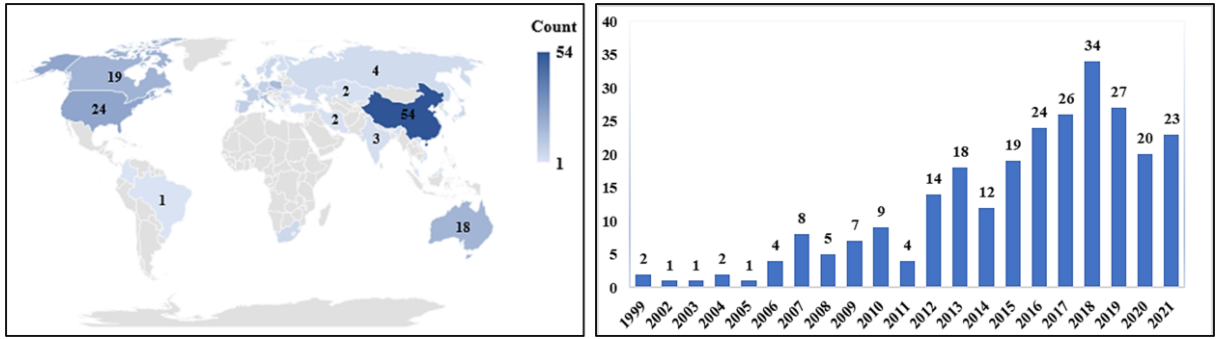

(a) By country          (b) By year

**Fig. 2.** Number of publications by country and year.

## 3 Mapping and mobility

Modern mining operations are assisted by computer-aided programs that use 3D imaging data collected from laser scanning to address industrial, laboratory and numerical research problems. This section reviews the fundamental factors associated with 3D mapping, and the level of mobility of scanning systems in underground mines with respect to their mode, methods and ease of data collection.

*3.1 Influence of sensor and environmental characteristics on 3D mapping*

Laser scanning uses the principle of LiDAR, where the time- or phase-differences between the emitted and reflected rays are measured to determine the distances between the sensor and target objects. Together with extrinsic parameters of sensors, such as position and orientation, measured ranges can be converted into a 3D point cloud, which represents the digitally scanned environment. The accuracy of the generated map and measured coordinates depends on factors such as the specifications of the LiDAR sensor and surface properties of the scanned environment.

*Sensor characteristics:* Prominent sensor-related factors leading to mapping errors include sensor beam divergence, inertial measurement unit (IMU) bias (for MLS) and improper intrinsic and extrinsic calibration of sensors (Fig. 3-a) [10]. Sensor beam divergence refers to the gradual increase in the footprint of a laser pulse with distance due to the solid conal angle projected by a laser beam (Fig. 3-ai). As a consequence of beam divergence, distant measurements tend to increasingly be affected by error due to a large uncertainty region in measuring a single point. To limit the error from beam divergence, a threshold cut-off distance is set, the maximum measurable range, beyond which readings are ignored. Where that range is set depends on the intensity of laser pulses, operational wavelength and signal attenuation over time [11]. Before using laser scanning for any application in an underground mine, characteristics of the scanner must be evaluated such as its range precision, beam divergence and minimum mappable unit (resolution/discernability) through suitable benchmarking tests for efficient mapping [1,12]. Such tests help to investigate the quality of data produced by the scanner and provide insights into the processing required to remove errors or inaccuracies from point clouds.

Beyond this, in mobile mapping, a primary source of error—besides odometry measurement errors caused by cycle slips—comes from inertial sensors, such as IMUs. Inertial sensors inherently accumulate drift over time, leading to a substantial difference in the measured and actual locations of sensors during mapping [3] (Fig. 3-aii). When a noisy output signal from an IMU is integrated, for instance integrating angular rate signals to determine angles, the integration drifts over time due to noise. The drift is referred to as a random walk as the integration seems to take random steps from one sample to the next. Angle random walk and velocity random walk are two main types of random walks in inertial sensors which apply to gyroscope and accelerometer measurements, respectively [13]. Orthogonality errors that consist of cross-axis sensitivity and misalignment, are also often encountered which impacts IMU measurement capabilities. Sensors mounted to an IMU need to be perfectly orthogonal. The cross-axis sensitivity error originates when a particular sensor axis responds to an input that is orthogonal to the sensing direction. Similarly, misalignment errors appear when the internal sensing axes do not align with the marked axes on the IMU case. In such cases, extensive equipment calibration between the sensors such as LiDAR or camera with IMU is required to overcome factory-calibrated misalignment errors [13–15]. Other sources of error in laser sensors and IMU include bias temperature sensitivity, scale factor and random error. The bias is a constant offset of the output value from the input value. When a sensor is started up, there is an initial bias present that may fluctuate based on thermal, mechanical and electrical variations [13]. Also, the bias drifts over time during operation at a constant temperature. Since the sensor operates in a range of temperatures, the biases may respond differently to each temperature that must be calibrated (Fig. 3-aiii). The scale factor is a ratio of output to the input over the measurement range. It varies with time and should be calibrated accurately before using the sensor [15]. Random errors exist inherently in the system and affect the precision of the measurements in LiDAR and IMU [13–15]. Improper intrinsic and extrinsic calibration of IMU and LiDAR sensors is also a major factor that results in distorted 3D maps. The extrinsic configuration of sensors can be disturbed by sudden sensor jerks, caused by the uneven terrain in underground mines (Fig. 3-aiv) [16]. Although not studied/compared extensively, the choice of scanning platforms such as backpack, helmet-mounted, handheld or vehicle-mounted laser scanners also impacts the point cloud characteristics (such as density, point spacing and accuracy) as speed, movements (lateral, rotational, static, vibrational) and calibration of sensors vary [2,8] (Fig. 3-av).

*Environmental influence:* Environmental factors such as the ambient illumination level can cause the laser detector to saturate, thereby affecting its sensitivity to precisely detect the laser return pulse and thus inducing errors in the measured ranges [17]. Similarly, water vapour or moisture tends to attenuate laser pulses by absorbing near-infrared (NIR) wavelengths (900–950 nm), which are typically used in laser scanners (Fig. 3-bi). Additionally, dusty conditions at mine sites contribute to noise as they obstruct laser rays (Fig. 3-bii). Such noise can be removed in post-processing steps using filters such as statistical outlier removal, $k$ nearest neighbour or connected component. The number of points captured on the surface in an underground mine depends on the material's ability to absorb NIR wavelengths, while the type of reflection depends on the surface roughness. For instance, shiny or mirror-like surfaces often lead to multipath due to specular reflection that prolongs the ray reception time, leading to inaccurate ranges (Fig. 3-biii). Scanning results are usually optimal when sufficiently Lambertian surfaces (surfaces causing diffused reflectance) are present in the environment as laser incidence angles tend to impact the precision of generated point cloud [18,19]. Mapping accuracy in a large area also depends on the structural complexity of the environment;

sufficiently distinct features such as planes, lines, edges or markers are essential for automated interframe stitching [20]. Underground mines, particularly coal mines, are highly symmetrical and feature-deficient since their walls are often coated with shotcrete or mesh. As a result, it is difficult to identify unique tie-points, resulting in inaccuracies due to false laser scan matches [3](Fig. 3-biv). A mapping solution is, therefore, better put to use in rock/metal mines with structured environments and Lambertian rock properties, as opposed to underground coal mines with highly repetitive, unstructured environments and light-absorptive coal properties. Beyond this, dynamic movements in the environment, e.g. of vehicles or machinery, or changes in the operational area also crucially affect 3D mapping as interframe matching is impacted by differences between scenes [18,21]. For example, as shown in Fig. 3-bv, during an active scan, a change in the ventilation door status induced a mapping error.

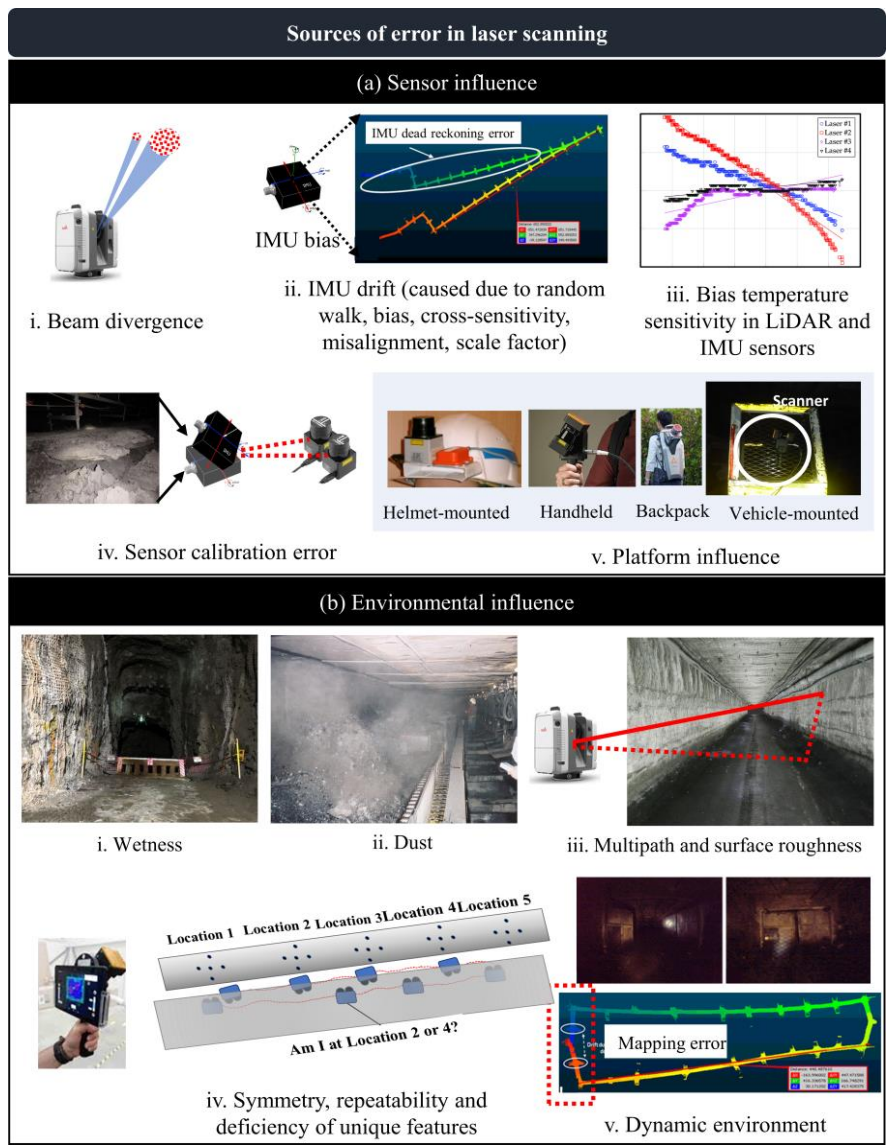

**Fig. 3.** Major sources of mapping errors through laser scanning in underground mining environments. Beam divergence [10], IMU error [13], sensor bias [13], calibration error [16] and platform influence are primary sensor related sources of error [2,8]. Whereas lighting condition [17], water, dust [22], multipath [1], surface roughness [19], and structural complexity [18,21] are environment related sources of error.

### 3.2 Degree of mobility in underground mines

Using TLS in an underground mine involves successive scanning from a fixed point followed by co-registration of point cloud scans in post-processing, which is inefficient when mapping large areas of mine tunnels [23]. Moreover, the co-registration process necessitates significant overlap between subsequent scans to achieve coherent stitching of point clouds, which is mostly done manually by identifying distinct fiducial marks [24] or through automated coregistration algorithms such as iterative closest point (ICP) or normal distribution transform (NDT) [25,26]. Key drawbacks of TLS are the arduous instrumental setup, difficulty in carrying around equipment, limited coverage area, presence of blind spots (absence of line-of-sight) and intensive processing for scan coregistration. In contrast, MLS in underground mines uses the SLAM principle to enable mobility. SLAM is a computational approach to determining the exterior orientation parameters (mainly position and orientation) of the scanning sensor while simultaneously building a map of the environment. Since the position of the sensor and surrounding map correlate, the solution narrows down the work required in an optimisation process [27]. Over the last decade, SLAM has been developed to suit indoor, underwater and underground field applications. Since various studies already presented systematic reviews of SLAM algorithms [28,29], we limit the review content here to briefly discussing application cases in underground mining.

#### 3.2.1 Algorithmic improvement

Researchers have made software and hardware modifications to facilitate the development of robust SLAM solutions for different environmental scenarios that may include feature deficiency, symmetry, repeatability and/or dynamic objects. Conventional SLAM algorithms used a probabilistic approach employing Bayesian filters to determine the pose of the scanner and the surrounding map. Kalman filtering (KF) formed the basis of such algorithms and was based on the assumption of linearity, i.e. the motion of the system described by linear equations [30,31]. However, in the real world, the motion of a system is seldom linear. Hence, the algorithm was extended for non-linear cases, which came to be known as the extended Kalman filter (EKF) [32]. KF and EKF filters were relatively slow when estimating high-dimensional maps due to the complexity of the equations and state matrix, and so dual algorithms were proposed in the form of the information filter and extended information filter, which resulted in the achievement of a sparse-state matrix for fast computation [33]. For real-time operations and to avoid extensive dependence on Gaussian filters, non-parametric filters such as the histogram filter [34] and particle filter [35] were used in SLAM. FastSLAM was a non-parametric filter-based SLAM that used a Rao-Blackwellised particle filter [35]. Yet, the above algorithms were vulnerable to accumulated sensor noise, which often result in mapping drifts [3,36]. These were reduced in later research by introducing optimisation-based SLAM, where a previously seen place is recognised through feature matching, i.e. loop closure, to correct the trajectory and map. The optimisation-based SLAM approaches that have been developed include graph-based [37], particle swarm [38] and LiDAR odometry and mapping (LOAM) [39,40].

Dissanayake et al. [32] reviewed SLAM algorithms and found that a conventional Bayesian approach such as EKF did not provide reliable estimates in larger scenarios, but when optimised, tended to generate better outcomes. The main limitation of optimisation-based

SLAM was highlighted as its poor results with false loop-closure detection, due to feature deficiency or high symmetry. To overcome this issue, researchers have suggested that rather than using all loop closures, it is acceptable to ignore a few that may corrupt results [41]. Recently, the focus has been on using deep learning-based semantic SLAM to improve loop-closure detection, as most of the encountered objects (common landmarks such as edges, poles or furniture) in indoor or open environments are known, meaning it is straightforward to prepare training data [42,43]. Use of deep learning also adds robustness in the presence of dynamic objects in the surroundings [29]. A comprehensive discussion on the past, present and future of SLAM was presented by Cadena et al. [28].

*3.2.2 Hardware improvement*

Hardware modifications to improve SLAM include adding sensor modalities such as an IMU, monocular camera, stereo camera, LiDAR, RADAR and SONAR, to achieve better localisation and mapping. This also ensures data ghosting can be avoided when one of the sensors fails to operate. Different combinations of sensors were tested in SLAM solutions such as LiDAR-only [44], IMU+LiDAR [45], camera-only [46], IMU+camera [47], IMU+camera+LiDAR [48] and IMU+camera+LiDAR+RADAR [49]. Mendes et al. [42] evaluated LiDAR-only SLAM with ICP and pose graph optimisation, observing that the SLAM solutions often converged to local minima instead of global, causing scans to drift over time in the absence of geometric features. Drift was resolved over a small scale (<1 km) by using an IMU sensor, which helped estimate the state (position and orientation) of the sensor by continually measuring the inertial vectors. Most current mobile laser scanners combine IMU and LiDAR to avoid excessively depending on geometric primitives in the environment [45]. At a large scale, IMUs inherently accumulate minute errors over time. Thus, accurate loop closures become essential to reduce dead reckoning errors induced by IMU. Adding data modalities through sensors such as optical, depth and multi-spectral ones on top of LiDAR improves results through better detection of loop closures [48,51–53].

*3.2.3 Practices in underground mines*

MLS systems are considered relatively new to the mining sector, but their impact on how we can collect useful information for critical mining applications is already being seen. The prominent off-the-shelf MLS systems currently in use are shown in Fig. 4. A large-scale presentation and comparison of multi-sensor versus single-sensor SLAM for mapping were offered by Jacobson et al. [22]. Two underground hard rock mines in Australia, Queensland (12.3 km mapping length) and New South Wales (32.3 km), were mapped using laser-only, camera-only and laser+camera+odometer+IMU sensors. A sensor positioning error of 0.68 m was observed for fused sensors in the Queensland mine dataset, which was comparatively lower than for the laser-only (0.94 m) and camera-only (4.49 m) SLAM. Similarly, for the New South Wales dataset, errors of 122.4 m, 1.38 m and 1.32 m were observed for the laser-only, camera-only and fused sensors, respectively, with superior results exhibited by fused sensors [22]. A laser-only solution is not sufficient for accurate mapping in underground mines; fusion with other sensors is necessary for a more robust solution. A comparative evaluation of TLS, MLS and structure from motion (SfM) for mapping voids in underground mines and tunnels was presented by Wong et al. [4]. The test area consisted of an unstructured (feature-deficient and with unknown geometry) corridor, structured corridor and unstructured intersection. The results showed greater errors with MLS and SfM for the unstructured corridor and unstructured intersection than for the structured corridor environment.

Most of the studies evaluate the quality of point clouds rather than comparing the data collection time using TLS vs. MLS for a given scale. Multiple factors influence the data collection time for terrestrial or mobile scanning. In TLS, equipment transport, setup, and the number of stationary data collection points are time-controlling factors that may vary substantially depending on the user's expertise. Whereas for MLS, factors such as speed of travel and number of loops decide data collection time. In a study, Neumann et al. [54] investigated two methods of mapping, (1) stop-and-go and (2) continuous scanning, by mapping 2.4 km of an underground tunnel [54]. The researchers concluded that continuous scanning methods hold advantages over the stop-and-go technique in terms of the data collection speed and processing time. A time comparison showed that continuous scanning can be achieved in nearly one-quarter of the time taken by the stop and go technique at a given scale. Further to this, a large-scale demonstration of mapping in underground mines was presented by Zlot and Bosse [55], where a 3D model of over 17 km was developed for a copper and gold mine in New South Wales, Australia. The generated 3D model was found to be a suitable approximation of the designed mine plan. To evaluate the efficacy of the MLS system (Zeb1 and ZebRevo laser scanners), a comprehensive set of experiments were performed outdoors, inside a building, inside a multistep helical tower and in an underground mine [2]. The observed absolute mapping errors were 9 cm for small-scale (<100 m) and 70 cm for large-scale (660 m) outdoor maps, 7.4 cm for the indoor scan, 5 cm for the helical tower and 60 cm for a 100-m scan of the complex underground mine. Observations revealed that whenever the scan time exceeded 25 minutes, mapping drift resulted with an observable bend in a long, straight feature-deficient corridor [2].

Most mine sites either use the MLS system in handheld mode or mount it on a scanning vehicle. The use of a fully autonomous mapping system is still rare in underground mines. Few studies have conducted experiments with a fully autonomous aerial system for mapping [56], localisation [57], exploration [58] or search and rescue [59]. Kim and Choi [60] demonstrated the use of an autonomous MLS system in an underground tunnel for localisation and 3D mapping through inter-scan feature matching. The experiment was performed over a small scale (25 m), and a root mean square error of 0.05 $m^2$ was achieved when the mapped tunnel section was compared to a manual survey. Meanwhile, when MLS was used in a dusty underground environment, substantial errors appeared in localisation and mapping for large scenarios [56].

As discussed above, developing an accurate 3D map over a large scale in an underground mine remains challenging due to factors such as (1) accumulated inertial sensor bias causing dead reckoning error, (2) false loop-closure detection due to a featureless or highly symmetric/repeatable environment, (3) a dynamic environment, (4) feature deficiency caused by a low scan rate of the equipment and/or (5) loss of data due to a sensor fault or low-resolution data due to a high travelling speed. Consequently, though the development of a robust solution that is intrinsically safe and applicable to all kinds of mine sites (metal and coal) is highly desirable, no such feat has yet been achieved.

A key issue to consider is how the walls and roof of underground mines are usually covered with shotcrete, thereby reducing the number of distinct features on the surface, meaning SLAM algorithms frequently encounter false loop-closures [61]. Although researchers have made significant contributions to the development of SLAM for large-scale underground mapping [2,55,62], there are still insufficient use cases or data to validate the robustness, precision and repeatability of the results under varying mining conditions and at various scales. Moreover, deep learning-based semantic SLAM, which is the current state-of-the-art, does not apply to underground mines as no corresponding semantically labelled training datasets are available. Moving forward, establishing a repository containing generally encountered underground objects, such as signboards, wired mesh or roof bolts, would be useful to boost MLS for monitoring applications. Beyond this, to facilitate the wider applicability of MLS in sensitive mining areas with potential gaseous emissions, such as in a coal mine, we must develop intrinsically safe scanners or flameproof enclosures that can host laser scanners and other multi-modal sensor systems [63].

| Name | System name | Base scanner \| Data rate \| Range |
|---|---|---|
| 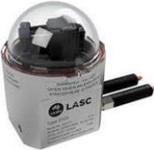 | CSIRO ExScan 3D | Hokuyo UTM-30LX \| 43,000 pts/s \| 30 m |
| 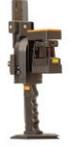 | GeoSLAM Zeb-Revo | Hokuyo UTM-30LX \| 43,000 pts/s \| 30 m |
| 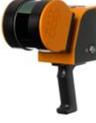 | GeoSLAM Zeb-Horizon | Velodyne Puck VLP-16 \| 300,000 pts/s \| 100 m |
| 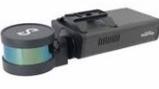 | Emesent Hovermap | Velodyne Puck VLP-16 \| 300,000 pts/s \| 100 m |
| 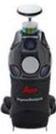 | Leica Pegasus | Leica laser scanner \| 600,000 pts/s \| 200 m |
| 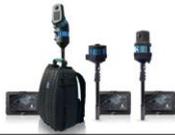 | Gexcel Heron | Velodyne Puck VLP-16 \| 300,000 pts/s \| 100 m |

**Fig. 4.** Off-the-shelf mobile laser scanning (MLS) systems available for mapping underground mines.

# 4 Data coherence/registration in underground mines

Primary applications in underground mines, such as localisation, convergence monitoring, structure mapping, change detection and deformation tracking, require either globally referenced point clouds (georeferencing) or mutually aligned multi-temporal point clouds (coregistration) for successful implementation. In underground mines, coherent multi-spatiotemporal point clouds are retrieved using a registration process that mainly entails georeferenced and coherent single-time data and coregistered multi-temporal data [18]. Single-time coherent data involve dead reckoning and accurate frame-by-frame coregistration to generate georeferenced maps, while multi-temporal coregistered data stem from large-scale feature matching to align multiple point clouds. Direct georeferencing, i.e. attachment of a global coordinate system (usually earth-fixed) of point clouds, is difficult in underground mines due to the absence of a spatial positioning framework. Instead, point clouds in underground mines are indirectly georeferenced by transferring datum from a GNSS-aided environment through a total station survey using ground control tags (GCTs) [64]. In the process, a few easily identifiable GCTs are installed on the surface, which are then tagged with a global 3D coordinate by demarking them using the total station equipment through triangulation. Subsequently, these tags are identified in the point cloud, either manually or through automated algorithms, for conversion from the local coordinate system to the global coordinate system [65].

Though georeferencing aligns multi-temporal point clouds, there are often shifts in the alignment as only a few discrete points are used in the process, and manual tagging errors may also arise. Therefore, fine refinement is necessary through the coregistration of point clouds. In coregistration, one of the scans is considered the base, from which other scans are aligned through rotation, translation and optional scaling. Coregistration of point clouds can either be done manually by identifying common features between point clouds or through automated algorithms, such as the commonly used iterative closest point (ICP) or normal distribution transform (NDT), which finds the optimal transformation between point clouds [18]. ICP coregisters by minimising the sum of squared errors between points in clouds, while NDT matches two point clouds through Newton's algorithms by dividing them into voxels and modelling the points' Gaussian distribution [25,66].

Despite several variants of ICP and NDT having been established, for all, coregistration in ICP depends on the initial point cloud alignment, and there are often poor convergence rates for NDT-related algorithms [67]. Therefore, to achieve efficiency, faster convergence and accurate coregistration, methods based on point cloud descriptors were developed. These involve identifying key points in the cloud using unique point descriptors such as the fast point feature histogram (FPFH), rotation invariant feature transform (RIFT), scale-invariant feature transform, regional curvature, etc., which are then collocated in point clouds for alignment [67–71]. Though traditionally, the computation of descriptors was expensive and consumed a lot of memory, current state-of-the-art coregistration algorithms exploit semantic registration, i.e. deep learning on commonly encountered objects, when aligning point clouds in indoor environments [67,72,73]. Dong et al. [67] provided several large-scale benchmark datasets of a subway station (560m), railway tracks (1200m), mountains (360 m), forest (87 m) and underground excavation (790 m), for use when testing and investigating deep learning-based point cloud registration. To date, however, there are limited semantically labelled datasets for underground mines, which makes the GCTs depicted in Fig. 5 the primary choice for automated georeferencing and coregistration.

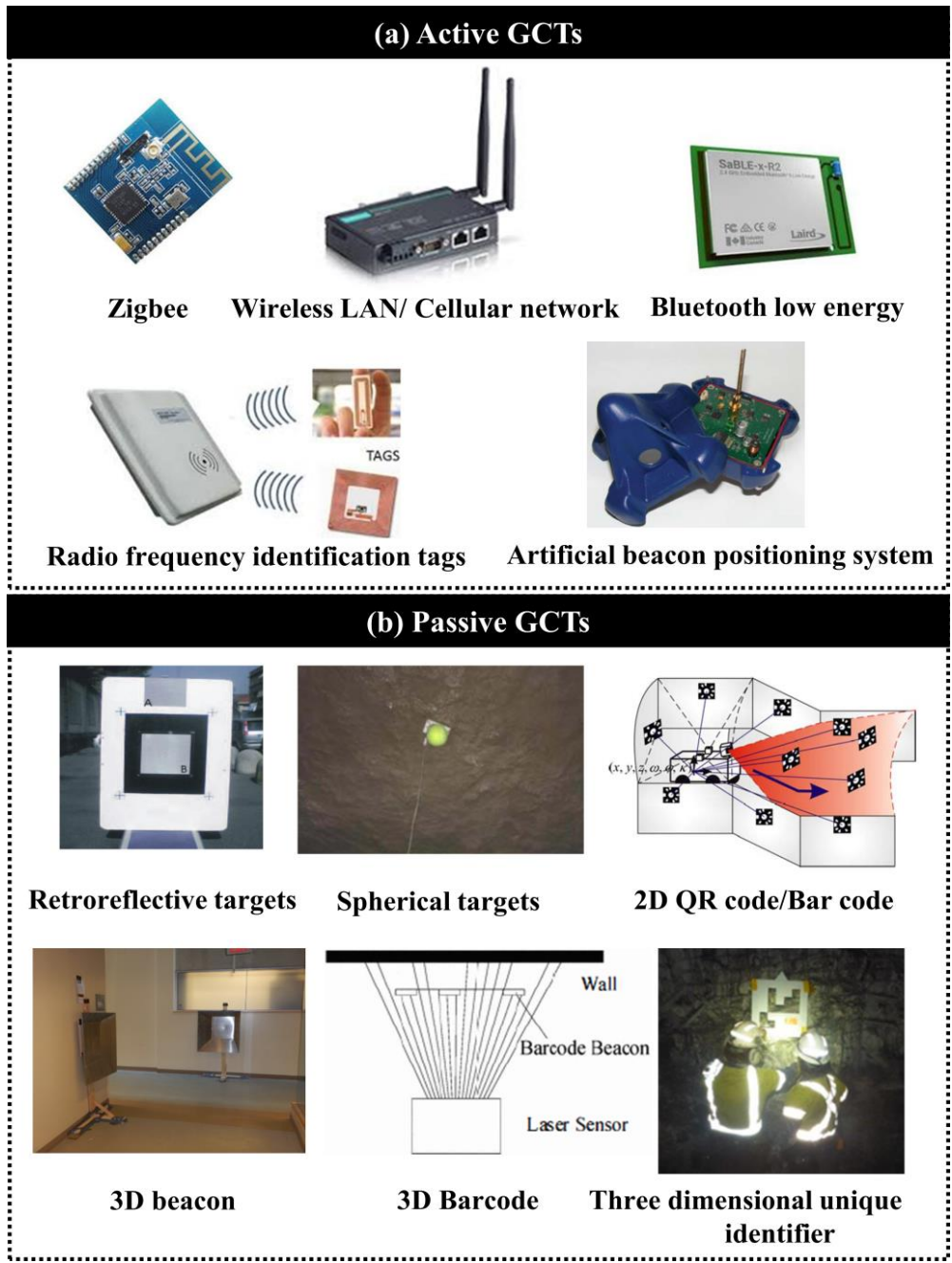

**Fig. 5.** Active and passive ground control tags (GCTs) available to improve mapping solutions in underground mines (subfigures adopted from [18,74–78]).

*4.1 Active GCTs to improve registration*

Registration of point clouds is achieved using GCTs by estimating the state (position and orientation) of the scanning device. With a known position and orientation of the sensor, multi-temporal point clouds are automatically georeferenced, meaning they are then easily refined using coregistration algorithms thanks to the initial alignment achieved. Most research on active GCTs involved radio frequency identification (RFID) tags or wireless sensor networks, while a few studies tried using light-emitting diodes (LEDs). Lavigne et al. [79] demonstrated the use of RFID as a GCT by creating a large-scale 2D map of an underground tunnel (5 km) and mine (2.4 km). The study divided the environment into subsets by detecting RFID, to determine the poses of the scanning platform locally and globally using graphs. Yet, since the study assumed there was a two-dimensional environment, mapping errors arose from changes in the inclination. A further investigation into the required number and distribution of GCTs revealed that mapping errors within a fixed coverage area remain consistent irrespective of the number of GCTs used [79]. However, the coherency of the generated map improves when more GCTs are used due to the uniform distribution of error over the area. The accuracy of RFID measurement—and, in turn, mapping—depends on multipath and cycle ambiguity [79,80]. Error in inter-frame registration can be avoided by using 3D laser scanners instead of 2D, which enables better scan matching using key features in the environment.

Motroni et al. [81] reviewed the localisation methods that employ RFID. The study suggested that using multiple sensors including proprioceptive sensors such as an IMU, tilt or odometer, as well as exteroceptive sensors such as laser scanners or cameras, together with RFID, improves sensor positioning and mapping while making the solution robust to data ghosting or corruption. A further study on RFID transmission power levels showed that location tracking is more precise with reduced reader transmission power, but the recognition rate also decreases [82]. Beyond these, studies investigated LEDs as an alternative to RFID, as recognition of RFID is affected by radio or electromagnetic interference and has a longer response time and low precision. In contrast, LEDs offer several advantages such as a longer life expectancy, higher tolerance of environmental hazards, low power consumption, lower heat dissipation, low costs and comparable accuracy to RFID [83,84]. However, LEDs are rarely employed in underground mines for localisation due to the dusty conditions, which affect their recognition. Similar to RFID, LEDs require a direct line of sight to the reader and recognition is limited to only short ranges, usually of less than 5 m.

In recent years, the focus has shifted towards wireless sensor networks (WSN) for underground mines, to enhance the efficacy of monitoring systems [78]. Muduli et al. [85] reviewed the suitability of wireless sensors in underground mining contexts and concluded that WSNs, when combined with monitoring sensors, can be used to monitor the mine gases, temperature, humidity, air pressure, dust, fires, explosions, roof falls, mine personnel and equipment. Commonly used wireless sensors for localisation and mapping include ultrawideband (UWB), Zigbee, wireless local area network (WLAN), cellular and Bluetooth low energy (BLE) [74] (Fig. 5a). Among the available wireless sensors, UWB provides the most accurate localisation (sub-centimetre accuracy), but it is expensive and affected by interference from metallic objects. For other active GCTs, the accuracy mostly remains in centimetres [85].

*4.2*  Passive GCTs to improve registration

Active GCTs require power to operate, making them expensive and limiting their use in safety-sensitive underground environments like coal mines due to fire-related hazards. Hence, intrinsically safe passive GCTs were explored, using retroreflective materials, since they exhibit higher-intensity values against the background structures in the point cloud [77,86,87]. An approach relying on intensity is not optimal for underground mines as reflectivity attenuates over time due to dust, humidity and corrosion. Certain studies focused on using barcodes and QR codes as GCTs, to obtain coherent data in GNSS-denied environments by recognising GCTs through a complimentary optical camera sensor [88–91]. Yet, the recognition of barcodes and QR codes using passive optical sensors depends on lighting conditions and is unsuitable for sub-optimally lit underground spaces [18]. Therefore, geometric GCTs, which are recognisable directly from the 3D point cloud scans, without the need for a complimentary camera sensor, are the most viable option for data registration [75].

Shi et al. [92] evaluated geometric GCTs of varying shapes and sizes for easy identification in point clouds during registration. Most of the objects were symmetrical but geometrically different from the surrounding objects/structures in the environment. For instance, spheres are commonly used objects in the data-registration process [93,94]. Recognition of a sphere is simple and thus accurate even in a low-resolution point cloud (5–10 cm point spacing) through precise location estimation of the centre based on surface reconstruction. However, symmetrical and similar GCTs, such as multiple planar panels or spheres, are virtually indistinguishable and thus do not contribute to the scanned environment in terms of marking unique locations. In most cases, field notes detailing the local environment of installed symmetrical GCTs, followed by their manual recognition and coordinate annotation in the point cloud, are essential for a successful operation [18]. Yet, such approaches are cumbersome in the routine large-scale scanning of underground mines, which involves regular registration of multi-temporal point clouds.

To improve on this, the operation can be simplified and automated using unique GCTs that can be differentiated among themselves. Accordingly, the need for unique identities in GCTs led to the development of a 3D barcode beacon for underground mining, which could be recognised in the 3D point cloud [89]. The process involved decoding surface variations of 3D barcodes in the point cloud for unique identification. However, 3D barcodes were hard to construct and rarely used due to detection concerns when surface variations in the environment were large and the point cloud was noisy. Simela et al. [76] developed a new 3D geometric beacon and proposed a workflow for automated georeferencing of point clouds in underground mines. The concept was demonstrated by placing surveyed GCTs (through the total station) opposite each other on the walls of the indoor environment and then decoding the target in point clouds for correspondence. The major limitations of the GCTs used were their large size, complex construction geometry and the subjective recognition of edges for decoding in a noisy point cloud.

Elsewhere in the literature, Singh et al. [18] developed 3D GCTs termed three-dimensional unique identifiers (3DUIDs) containing unique embedded patterns of slots (or gaps) and solid fixtures (refer to Fig. 5b). The proposed GCTs were inexpensive, simple to construct, easy to install and easily decodable in a 3D point cloud. The study compared 3DUID-based direct georeferencing and coregistration of point clouds with conventional ICP and NDT registration algorithms and achieved a comparable accuracy, to within half a metre, for a kilometre-long scan of an underground coal mine. Further existing research also suggested that placing GCTs along the longer section of a tunnel reduces dead reckoning issues in visually similar and symmetrical environments.

*4.3    Challenges and scope for improvement*

In underground mines, the scale of a scan decides the success of the application using coregistration. For a small-scale point cloud, collected within an allowable time limit before dead reckoning errors occur, coregistration can be achieved with high accuracy using any conventional algorithm, though the range accuracy of the scanner given in the specification sheet may impact the results [2,8,95]. Fully automated large-scale (>1 km) applications in underground mines require a spatial frame of reference to correct the distortion caused by dead reckoning errors. The reference frame can either be global or local and can be obtained by employing active or passive GCTs. The types of GCTs used depend on the condition of the underground environment. Underground metalliferous mines present fewer challenges than coal mines in terms of the complexity and intrinsic safety of sensing tools [18]. Hence, active sensors/GCTs can be employed in hard rock mines as there are fewer electrical power constraints than in coal mines. Moreover, hard rock mines usually have more exposed structures, providing sufficient visible distinguishable features for data registration than in coal mines.

In its current state, MLS alone fails to provide accurate, large-scale mapping solutions for an underground coal mine due to the high symmetry, repeatability and feature-deficiency of such a mine. Yet, together with active and passive GCTs, accurate monitoring solutions could be developed. Active GCTs offer several advantages such as real-time mapping, localisation and data transfer, but they are expensive and not intrinsically safe to operate in safety-sensitive coal mines. In contrast, passive GCTs are intrinsically safe and support real-time localisation and mapping. Today, both active and passive GCTs can produce coherent data under conditions of feature deficiency, symmetry, repeatability and GNSS signal absence. However, further research is needed to make active GCTs intrinsically safe to operate in coal mines. With that in mind, in future, a combination of unique 3D GCTs together with multi-imaging sensors, such as LiDAR, IMU, optical/multispectral/thermal cameras and RADAR, could be explored to support the development of better mapping solutions.

## 5    Geological and geotechnical applications

In this section, we discuss geological and geotechnical applications that have been influenced by the use of laser scanning.

*5.1    Lithological classification*

The lithological composition of a rock mass is an important parameter affecting how we compute its geological strength index and characterise the rock mass [96]. While multi-spectral and hyperspectral imaging are the preferred approaches for analysing lithology, their performance is subject to the lighting conditions in underground mines. In that context, several studies explored laser-scanning solutions to estimate the lithological composition of a rock mass by utilising geometrical point descriptors and intensity information [97,98]. Walton et al. [99] segregated mudstone and sandstone in a rock mass by defining several descriptors based on the intensity and surrounding geometry, which were then used to train a decision tree for classification. In a similar study, textures and local descriptors were used to map cherts in outcrops by training a support-vector machine classifier [100]. Cross-validation of the results against field inspections and reference pictures showed that TLS data can be efficiently exploited to map chert. The concept was further proven by Živec et al. [98], where the intensity difference captured from TLS was compared with results from X-ray diffraction and lithological profiling (by measuring the bed thickness) to differentiate marlstone from sandstone. The results showed notably different intensity values for the two rock types.

To date, however, in underground mines, limited-to-no studies employed laser scanning to identify the lithological composition of the rock mass. Looking ahead, further research using laser scanning for lithological classification could improve the rock mass characterisation process in underground mines with suboptimal lighting. Given the recent advances in multi-spectral LiDAR, further improvements in identifying the lithology are also expected [101,102]. However, multi-spectral LiDAR is expensive, bulky and presents several technical challenges associated with calibration and sampling of intensity values together with 3D measurements. Accordingly, it will take some time for the technology to mature for mobile 3D multi-spectral laser scanning in underground mines.

*5.2    Rock mass discontinuity mapping*

Discontinuities are planes of physical or chemical variation within a rock mass, which depend on the geological condition(s) under which the rock formed. Underground mines require extensive investigation of rock mass discontinuities as the heterogeneity and anisotropic features of these discontinuities are crucial to analysing hydraulic, mechanical and deformational behaviours of the rock mass, which influence its stability. The primary discontinuity characteristics used in stability analysis are the orientation (dip angle and direction), spacing, persistence (trace length), strength filling, roughness, lithology, seepage, number of sets and block size [12,103].

Conventionally, these rock mass characteristics were studied through field observation by mapping discontinuities using a compass and measurement tape. Recently, however, laser scanning-assisted, automated discontinuity mapping has been becoming a popular alternative as it is efficient, simple and accurate, while providing the opportunity to completely cover discontinuities over wide scales. Discontinuities can be detected from laser scans through manual, semi-automated and automated approaches. The manual approach relies on the expertise of individuals to identify discontinuity planes in 3D point clouds, which is analogous to placing a compass on a rock mass, but done digitally using a virtual compass [104,105]. In this method, a plane is fitted on the selected region to estimate the dip direction and dip angle of a discontinuity. The semi-automated approach to identifying a discontinuity in a region typically involves a plane-detection algorithm combining user-defined parameters such as the threshold point distance from planes, threshold plane orientation for a discontinuity set and number of points for plane fitting [106–108]. Gigli and Casagli [106] launched a semi-automated MATLAB tool "DiAna" that uses singular value decomposition to determine the best-fitted plane in a cuboidal region. The user is required to define the dimension of the cuboidal region and plane threshold to identify discontinuities. Similarly, discontinuity set extractor (DSE), a MATLAB software developed by Riquelme et al. [109], is a semi-automated open-source tool that uses plane detection followed by kernel density estimation to identify major discontinuity sets.

Most semi-automated approaches take user-based constraints into account, but defined parameters are not universally applicable, thus requiring trial and error to achieve the best performance. On the contrary, automated approaches identify discontinuities with minimal human-defined parameters, instead working as 'click and play' tools. However, automated approaches frequently cause inaccuracies due to false discontinuity identification. The automated approach takes the entire point cloud as its input and adaptively computes parameters based on point cloud characteristics, to classify discontinuities into individual sets. For instance, several studies employed octree [110], triangulation [111] or a multi-scale region [112] to automatically capture the geometry around a query point without explicitly defining the local region. The need to define the plane threshold was avoided by developing clustering-based approaches or region-growing approaches that used the obtained signatures to segregate discontinuity planes [103,113,114]. The benefits and limitations associated with each of the three approaches (manual, semi-automated and automated) are listed in Table 1, which lists the software and algorithms available in each category.

**Table 1.** Benefits and drawbacks of various approaches to identifying discontinuities in a 3D point cloud.

| Approach | Benefits | Drawbacks | Existing software/algorithms |
|---|---|---|---|
| Manual | • Highly accurate<br>• Highly reliable when done comprehensively<br>• Effectively captures discontinuity boundaries for trace length<br>• No definition of parameters is required<br>• Takes users' expertise into account | • Point cloud browsing is tedious<br>• Time-consuming<br>• Requires individual's expertise for discontinuity identification<br>• Might not give accurate measures if important discontinuity planes are missed | RocScience DIPs<br>CloudCompare virtual compass plugin |
| Semi-automated | • Quick results<br>• Threshold parameters constrain plane orientation in discontinuity set, thereby improving accuracy | • Requires extensive definition of parameters<br>• Point cloud characteristics, such as accuracy and resolution, affect results<br>• Misses important discontinuity planes due to manually defined parameters<br>• Same set of parameters may not work on different datasets | 3DEC<br>MoFrac<br>Maptek PointStudio<br>Sirovision<br>Coltop3D<br>Surpac<br>Shapemetrix 3D<br>Discontinuity set extractor<br>DiAna<br>Plane detect<br>CloudCompare facet plugin |
| | | • Discontinuity plane boundary ambiguity for trace mapping<br>• Multi-segment for the same discontinuity in the presence of surface variations | ISODATA clustering algorithm |
| Automated | • Works on the entire point cloud<br>• Selective number of universal parameters required<br>• Leads directly to final output<br>• No human intervention is required during processing | • Often time-consuming<br>• Occasionally identifies non-discontinuity planes<br>• Discontinuity plane boundary ambiguity for trace mapping<br>• Multi-segment for the same discontinuity plane in the presence of surface variations | SplitFX<br>Clustering on local point descriptors (CLPD)<br>Clustering by fast searches and finding density peaks (CFSPD)<br>Silhouette-based $k$-means clustering<br>Region growing |

Limited case studies have now been published on automated/semi-automated discontinuity mapping in underground mines [96,108]. Regardless of the approach taken, however, the fundamental concept of identifying a planar discontinuity remains the same. Thus, it is important to discuss the algorithmic development in discontinuity mapping, which can be applied with or without modifications to underground mines. In the mining industry, RocScience DIPS and Maptek PointStudio are among the widely used commercial software for discontinuity mapping from laser scans. These pieces of software are semi-automated and depend on the user's ability to map discontinuities in colourised point-cloud scans [12]. In an uncoloured point cloud, the browsing process to identify discontinuities is challenging and requires greater analytical expertise and skills of interpretation for precise results. In recent years, several algorithms were developed by the scientific community to promote the development of open-access tools for semi-automated and automated mapping of the discontinuities of a rock mass. Such tools often incorporated significant leaps in discontinuity characterisation algorithms involving clustering, region growing and deep learning (Fig. 6).

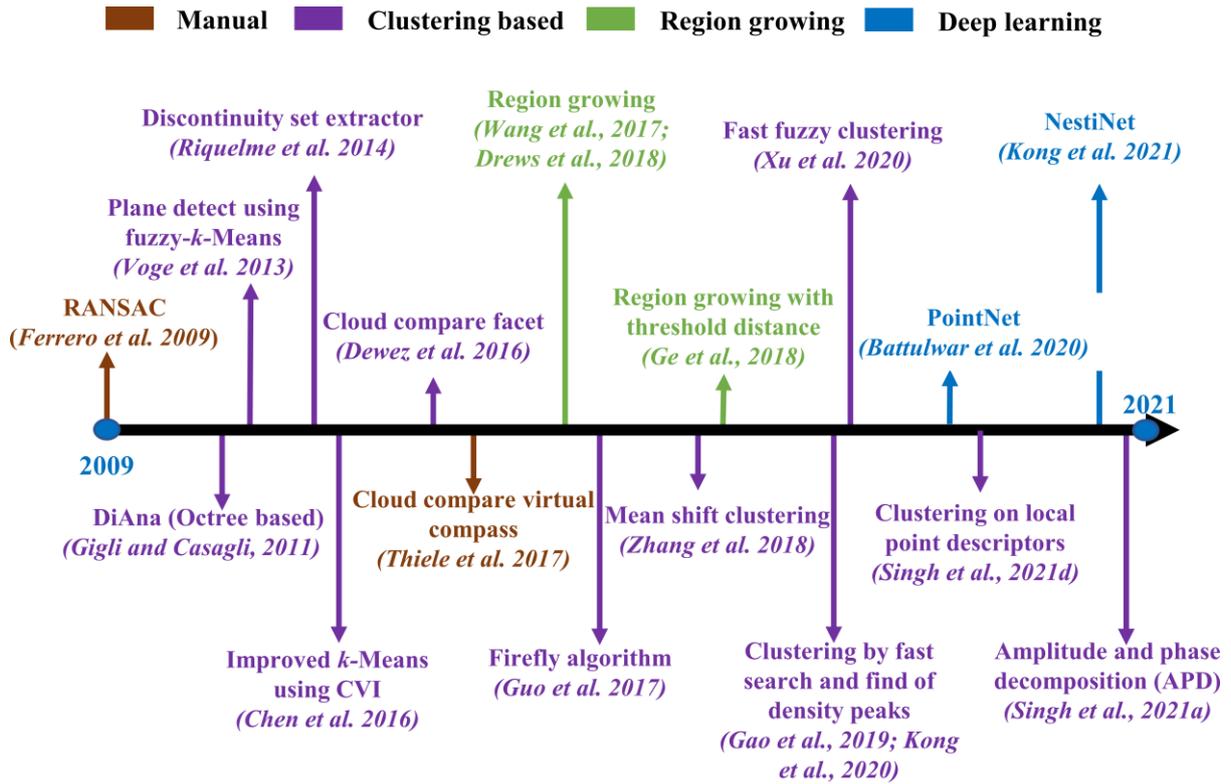

**Fig. 6.** Chronological development of discontinuity characterisation algorithms.

Semi-automated and automated algorithms in discontinuity mapping largely take data-clustering- or region-growing-based approaches. The workflows involved in the two approaches are presented in Fig. 7. Clustering approaches segment discontinuity planes by clustering point normal vectors or other point descriptors that describe local regions around the point. The approach requires only the number of clusters as the input, which can be automated by using cluster validity indexes (CVIs) such as Silhouette, DaviesBouldin or CalinskiHarabasz [115]. Chen et al. [116] applied *k*-means clustering to point normal vectors, derived from triangulated mesh, using Silhouette CVI, and then filtered discontinuity planes using a random sample consensus (RANSAC) plane-detection algorithm. From an algorithmic perspective, utilising a plane-detection algorithm before clustering is more sensible as points are constrained to discontinuity planes that facilitate better convergence of CVI, to identify an optimal number of clusters or discontinuity sets. Other methods to automatically identify the optimal number of clusters include the mean shift clustering algorithm (MSC) [117], genetic algorithm [110], clustering by fast searching and finding density peaks (CFSDP) [103,118], the histogram of point normals (HON) [12] and histogram of amplitude and phase [119]. Density-based approaches such as MSC, CFSDP and HON work on the density of point normal vectors' distribution, mainly their dip angle and dip direction, and converge towards the denser area. MSC and CFSDP utilise the angular distance between normal vectors, whereas HON runs a peak detection algorithm in a 2D histogram of point normals. After removing most non-discontinuity points through filtering, the number of clusters mainly depends on how accurately the normals are determined in the point cloud. Thus, laser scanner characteristics such as the range precision, minimum mappable unit and presence of noise play a crucial role in determining the success of the above methods.

Another important factor in identifying optimal clusters is selecting an optimal local region for robust estimation of normal vectors or point descriptors. Normal vectors tend to show high variability under the influence of noise or when the selected local region size is suboptimal. To master local support region selection, use of the octree structure was suggested, which divides the point cloud into voxels based on the point cloud density [110]. The process requires a user-defined minimum size of voxel as the input for voxelisation and then applies a plane-detection algorithm to estimate normal vectors. A genetic algorithm (firefly) was tested for its efficacy in identifying the optimal number of clusters based on their optimal convergence. However, the genetic algorithm was computationally expensive. Zhang et al. [111] suggested point-cloud triangulation followed by mean-shift clustering to avoid the need to select a local region when estimating the optimal cluster number. In recent research, Kong et al. [112] defined multiple local regions around the point to capture the surface geometry, a process that overcame the limitation of manual scale collection. The robustness of normal estimation and number-of-clusters identification was further enhanced by defining local point descriptors, such as the fast point feature histogram, eigenvalue descriptor and radial surface descriptor, to capture unique surface properties in terms of the point distribution, surface geometry and local surface variations, respectively [12]. The study also proposed an optimal radius of influence for point clouds with different point spacings through empirical testing and suggested point reprojection to remove local noise for robust normal estimation.

Region-growing uses the angular difference between normal vectors and a user-defined threshold criterion to identify whether a given point belongs to the same or a new region [114,120]. Hence, region-growing approaches are often affected by local variation in normal vectors that results in multiple segments for a single continuous discontinuity plane. The issue can be addressed to some extent by introducing additional user-defined constraints on the point distance from the fitted plane [121]. However, defining manual thresholds can hamper the segmentation process, as sometimes, the segments are either overestimated due to the inclusion of non-discontinuing points or underestimated due to the use of multiple segments for a single continuous regional plane. Consequently, the approach is unreliable for calculating the trace length and spacing of discontinuities, relying on a high-resolution, accurate point cloud for optimal performance.

The fundamental difference between cluster-based and region-growing algorithms is that the region-growing approach clusters segmented regions, while the cluster-based approach clusters individual points. After point clustering or region-based segmentation, both approaches require individual plane segmentation in a particular discontinuity set using density-based spatial clustering (DBSCAN) or Euclidean distance-based segmentation [23,121]. Then, a plane-fitting algorithm is applied, such as RANSAC, affine fit or principal component analysis, to obtain the dip angle and dip direction for Stereonet plotting. Few studies have investigated machine learning-based approaches to identify discontinuities in the point cloud [112,122,123]. In those that have, Kong et al. [112] modelled the point distribution through the Fisher vector distribution and then used a Nesti-net convolutional neural network [124] to identify the optimal scale for robust normal calculation. Finally, a fuzzy *k*-means algorithm was applied on normal vectors to cluster discontinuing sets. In a similar study, Battulwar et al. [123] evaluated a PointNet deep-learning algorithm to classify discontinuity points and then segregated them using density-based spatial clustering (DBSCAN). Yet, the method assumed that two discontinuity planes are always segregated by a distance after classification. A potential drawback of the approach is thus the difficulty in segmenting connected planes, along with how the authors ignored the possibility of a wedge-shaped discontinuity occurring.

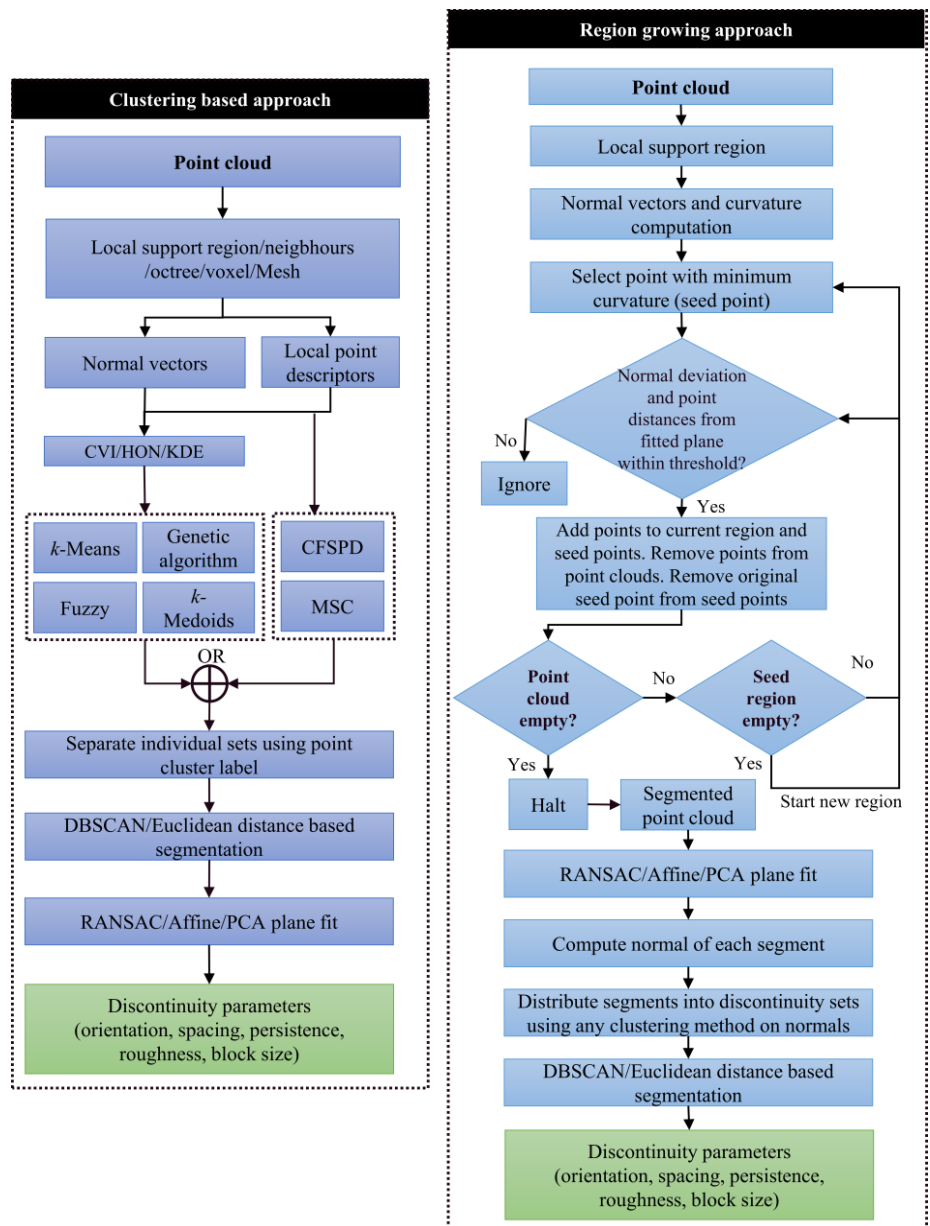

**Fig. 7.** Steps used in clustering-based and region-growing approaches for discontinuity identification in a 3D point cloud.

Battulwar et al. [125] reviewed conventional algorithms and found that previous algorithms were not sufficient to provide reliable discontinuity identification in a reasonable timeframe. Most conventional algorithms relied extensively on point normal vectors to identify discontinuity sets and experienced performance issues when there was noise or a low data resolution. Such issues could be resolved by using several local point descriptors at the expense of the processing time. Recent research avoided using normal vectors altogether and instead proposed the amplitude and phase decomposition (APD) of discontinuity points through fast

Fourier transform [108]. The method avoided pre-processing steps and led to the accurate mapping of discontinuities while reducing the processing time when compared to conventional algorithms.

Studies in underground environments have used manual discontinuity segmentation [126,127], commercial software such as SplitFX [62], Sirovision [128] and Maptek PointStudio [108,118], an open-source software discontinuity set extractor (DSE) [23] and algorithms such as CFSDP [118], region-growing [96] and the APD approach [119] to map discontinuities, where a gradual transition from manual to automated methods has been observed. The accurate identification of discontinuity planes in a 3D point cloud is essential to derive parameters such as the discontinuity trace length [129], spacing [130], roughness [103,109] and blocky-ness [112], which are required to analyse the stability of a rock mass based on its discrete fracture network. Beyond these, the orientation of discontinuities also helps to analyse the kinematic feasibility of wedge, toppling and planar failures in a rock mass [12] (Fig. 8).

Most underground case studies focused on mapping discontinuities on an exposed surface oriented in a particular direction. The most challenging aspect in the underground environment for discontinuity mapping is the enclosed nature of the rock mass, i.e. present on all the sides. For an enclosed area, segregating discontinuities becomes difficult as the mean orientation of the rock mass cannot be estimated unless it is manually segmented into the left/right wall and roof. However, such segmentation cannot be achieved for a more complex underground environment, as is particularly the case for stopes, which do not have a clear distinction between walls [131]. This specific scenario requires finetuning of parameters and trial-and-error if semi-automated/automated approaches are to succeed. Studies that mapped structural discontinuities in stopes utilised manual identification in point clouds, with limited-to-no studies employing automated/semi-automated methods [131,132]. Another challenge in underground mines is the need to separate natural discontinuity planes from other artificial planar structures. As an example, portions of walls may appear planar despite not actually being a discontinuity; separating such features requires manual intervention.

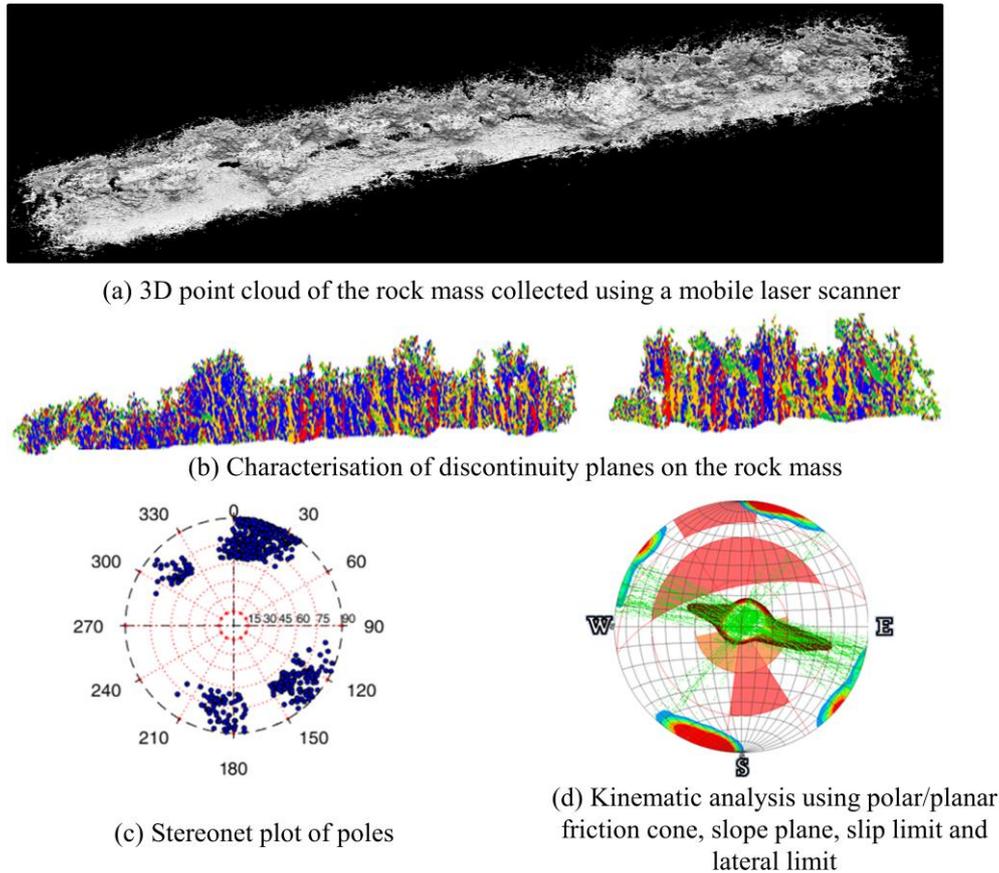

**Fig. 8.** Kinematic analysis of failure type though discontinuities identified from 3D point cloud (modified from Singh et al. [12]).

A major limitation of laser scanning is its susceptibility to blind spots beyond the sensor's line of sight, which lead to inaccuracies in tracing the discontinuity length and spacing and in performing kinematic analysis. Blind spots can be reduced by using MLS over TLS as a given scene is inherently captured from multiple views. A proper surface reconstruction technique may further reduce the effects of blind spots in discontinuity set characterisation, i.e. either by point resampling using triangulation or through spline surface fitting [133]. Discontinuity characterisation could be further improved by researching deep learning techniques as only limited studies in this regard are currently available [123]. The presence of an openly available dataset repository such as "Rockbench", consisting of scans collected from TLS, MLS and SfM in a range of operating environments, could prove useful in training and testing deep-learning models for discontinuity characterisation [134].

### 5.3 Rock mass characterisation

Rock mass characterisation is the process of quantifying a rock mass to explain possible ground behaviours that impact the design, planning and stability of engineering projects [135,136]. Quantification is carried out using indexes such as the rock tunnel quality index Q system, rock mass rating (RMR), and geological strength index (GSI), which depend extensively on the distribution and frequency of discontinuities mapped (Section 5.2). Q-system was specifically developed for tunnels and uses six parameters which include RQD, degree of joint alteration and clay filling, number of joint sets, joint roughness, water inflow and stress reduction factor. There are only limited studies which have used laser scanning to estimate the Q value [137,138]. Laser-scanning studies in underground mines have primarily used RMR or GSI for rock mass characterisation [113,127] while there are limited studies on the estimation of Q-values using laser scanning [137]. The RMR index uses six parameters: the uniaxial compressive strength (UCS) of rocks, rock quality designation (RQD) value, joint and bedding spacing (JS), joint condition, groundwater condition and orientation of discontinuities with respect to the opening axis [139]. Similarly, of these, the RQD, JS and orientation of discontinuities can be estimated confidently and accurately, while joint conditions (aperture, roughness, infilling and weathering) can be inferred partially using laser scanning [113,140]. Discontinuity characteristics (RQD, JS and orientation) are a critical component of RMR and Q-system, but estimates using conventional methods of manual in-situ measurement collection were not only subjective but also challenging in practice. This underpins the importance of laser scanning as an accurate and quick automated solution, though additional laboratory tests (for UCS) and in-situ visual observations (for assessing infilling, weathering and groundwater conditions) are still required for accurate RMR estimates.

In GSI, the original qualitative index proposed by Hoek and Brown [141] provided rough information about the rock mass strength. Later, the index was updated to incorporate quantification using information on rock block volumes and the joint surface conditions (roughness, weathering and infilling conditions) [142,143]. The block volume is estimated by joint properties such as spacing, orientation, persistence and the number of sets, all of which can be determined from a 3D point cloud using approaches given in Section 5.2. Recent underground studies based on laser scanning utilised discrete fracture networks created from joint properties to determine the rock block volume [144,145]. A discrete fracture network is usually created in proprietary software such as MoFrac (Mirarco Mining Innovation, Ontario, Canada) or 3DEC (ITASCA Consulting Group, Minneapolis, MN, USA). The joint surface condition, meanwhile, another important parameter in GSI, is obtained partially using laser scanning. Mah et al. [146] and Bao et al. [138], for instance, showed how the joint roughness can be estimated accurately using a 3D point cloud obtained from laser scanning. However, besides this, infilling and weathering of the rock mass have to be manually assessed in the field to obtain an accurate GSI value.

### 5.4 Change and deformation analyses

The purpose of change detection and deformation analysis is to ascertain variations in the geometric state of a scene in an underground mine over time. Both change detection and deformation analysis operate on multi-temporal point clouds, though they fundamentally differ in terms of quantification. The methodology for measuring change or deformation involves a pair of reference and transformed point-cloud scans, which are coherently registered (Section 3.2) The main types of change-detection algorithms are point-to-point (P2P), point-to-model (P2M) and model-to-model (M2M) comparison [1]. In the P2P method, the distance of a point in a temporal point cloud from the nearest point is computed in the reference scan. If the distance of a point is above a certain user-defined threshold, then it is assumed to have changed or deformed. The P2P approach requires point clouds with the same resolution for better correspondence and is unsuitable for point clouds from different sensors, i.e. multi-resolution. The P2M and M2M methods effectively overcome multi-resolution issues by reconstructing a point cloud through triangulation or local surface fitting. In P2M, the distances of points in the temporal point cloud from the nearest reconstructed surface are computed in the reference scan to detect changes, while in M2M, the temporal scan is again reconstructed, but this time, the distances of surfaces are computed instead of points [1,147].

In underground mines, change detection is used to estimate convergence and track local deformations in the environment to mitigate safety risks. The walls, roof and floor converge under the influence of horizontal and vertical stress fields, leading to phenomena like roof collapses or floor heaving. Timely management of these events is essential to avoid expensive downtime caused due to loss of production. Slaker and Westman [148] demonstrated the use of M2M comparison of TLS point clouds to quantify millimetre-level changes in the excavated coal area in an underground coal mine. In a drill and blast tunnel, Navarro et al. [149] used a laser-equipped navigation system to track deviations during rock mass excavation. The study installed track plates in the excavation area and then a 3D point cloud obtained from a laser scanner was used to track the drill deviation through multi-temporal scans, to verify whether the excavation was following the mine plan. In a similar study, multi-temporal point clouds were used to estimate excavation changes caused by redistribution of stresses in Creighton Mine in Sudbury, Canada [150] and Subtropolis Mine near Petersburg, OH, USA [151,152]. The point clouds were first registered using the ICP algorithm, then a P2P change-detection algorithm was applied to provide a quantitative estimate of changes [150].

Beyond those described, further studies also suggested analysing the obtained parameters, such as changes in the spatial location and volumetric quantity, to estimate the redistribution of stresses over time. In addition, a laser scanner was used as complimentary technology to ascertain the stress distribution leading to change in an underground limestone mine [153]. Multi-temporal observations revealed that TLS captured changes when seismic sensors recorded no events. Seismic sensors, being located at discrete spatial locations, can only record events when they occur within a certain defined vicinity of the installed sensors [153]. In such scenarios, laser scanners could provide relevant change information, and when combined with seismic sensors, stressmeters and extensometers could provide early warning signs for potential hazards.

A key issue in change detection is the selection of regions for coregistration. If the selected region is too large, then point cloud distortions may not be mitigated. Conversely, if the selected region is too small, then local movements in the area may be masked. Given this trade-off, most studies scanned well-established small areas in their change detection and deformation analysis [1,67]. In opposition to this, for large-scale routine assessment, the coregistration and change-detection accuracy depend on the ability to identify a previously scanned area. Change detection in underground mines often becomes more challenging when a section of a mine, most often the roof or floor, moves as a whole. In these scenarios, coregistration of multi-temporal point clouds results in false laser scan matches, as key matching points often lie in the moving/moved part, i.e. there are a lack of stable references. Such an issue could be resolved by establishing a spatial reference framework, either by using active or passive GCTs, so that multi-temporal point clouds could be automatically registered [18]. However, these GCTs would need to be installed in stable locations or surveyed routinely for accurate point-cloud registration.

In this area of study, Kukutsch et al. [154] monitored floor heave in an underground coal mine by collecting georeferenced point clouds over three years. Time-lapse scans indicated that the bottom of the seam displaced more than the top of the rib due to low floor strength, causing floor heave of up to one metre. Scans were coregistered using spherical targets and then a change-detection algorithm was applied to identify the floor heave rate. Currently, laser scanners are not used to relay continuous real-time change detection due to the lack of a data transfer framework and the high computation requirements. Thus, most of the studies employed post-processing of point cloud scans to detect changes [147,150]. Gálai and Benedek [155] attempted real-time change detection of moving objects using previously collected MLS data as a reference, comparing those with data from multi-beam rotating LiDAR. Similarly, Kromer et al. [156] utilised automated terrestrial laser scanning for near-real-time detection of rockfall and ground failure events. The data were collected at 30-minute intervals and processed almost in real-time using an onboard processing unit. However, no such studies can yet be found for underground mines. Laser scanning is often employed with complementary sensors such as extensometers, stressmeters, closuremeters and seismic sensors for comprehensive change and deformation analysis in the mining environment [157].

*5.5  Roadway clearance measurement*

Measurements of roadway clearance, overbreaks, underbreaks and water leakage are essential for safe operating conditions, optimal excavation and collision-free passage of vehicles. Mine operators are required to assess roadways routinely [158], which generally requires an absolute frame of reference. Pejić et al. [64] used a TLS to inspect the geometry of a 1260-m railway tunnel with the support of retroreflective GCTs for absolute referencing. Field validation of the TLS measurements with the ground truth, collected throughout the total tunnel length, led the researchers to conclude that the positional accuracy, when compared to empirical tunnel modelling, was around five times better, with a standard deviation of 2 cm. A similar study was conducted by Pinpin et al. [159], where a combination of TLS and selective targets was used to estimate the overbreak and underbreak in a railway tunnel. Laser scans collected at each location were registered using spherical GCTs. Since the design of the tunnel was known, the contour was compared to the collected scan to estimate the overbreak and underbreak locations and volumes, to support the collision-free passage of vehicles. Puente et al. [160], meanwhile, investigated an MLS system when used to obtain a cross-sectional profile of a roadway tunnel. An Optech Lynx mobile mapper system, equipped with GNSS, was used to survey a 200-m-long tunnel section, and roadway markings were used as references to extract cross-sections. The comparison of the profile with the ground truth revealed that the relative error in vertical clearance was mostly within 1%, with a maximal error of 4.5 cm.

Recently, Singh et al. [18] demonstrated a concept of multi-temporal roadway clearance tracking through GCTs, as presented in Fig. 9. The study developed a coregistration workflow and multi-temporal profile-matching concept to determine the convergence rate in roadway profiles. Field validation against ground measurements collected using a laser distometer exhibited a mean error of 5 cm. MLS was also used in room and pillar mining in an underground oil shale mine located in Estonia to measure cross-sectional areas of pillars and extracted volumes that were cross-validated against a TLS survey [8]. The errors in measured cross-sections and volumes between reference and MLS scans were 0.5% and 0.7%, respectively. The primary reason for errors was an incurred shift in MLS scans due to the dead reckoning error over an extended time. Nevertheless, the study showed that the results achieved were 10 times better than the discrepancy values allowed in contemporary Estonian mining regulations. Collectively, these studies attest that laser scanning is currently the most appropriate technology to accurately measure roadway clearance over a large scale. Choosing between TLS and MLS is dictated by the resolution and accuracy requirements. For instance, with measurements at a sub-centimetre level, such as for clearance, it is most appropriate to use TLS, while for an application that requires lower spatial resolution and accuracy, such as measuring overbreaks and underbreaks, it is better to employ MLS.

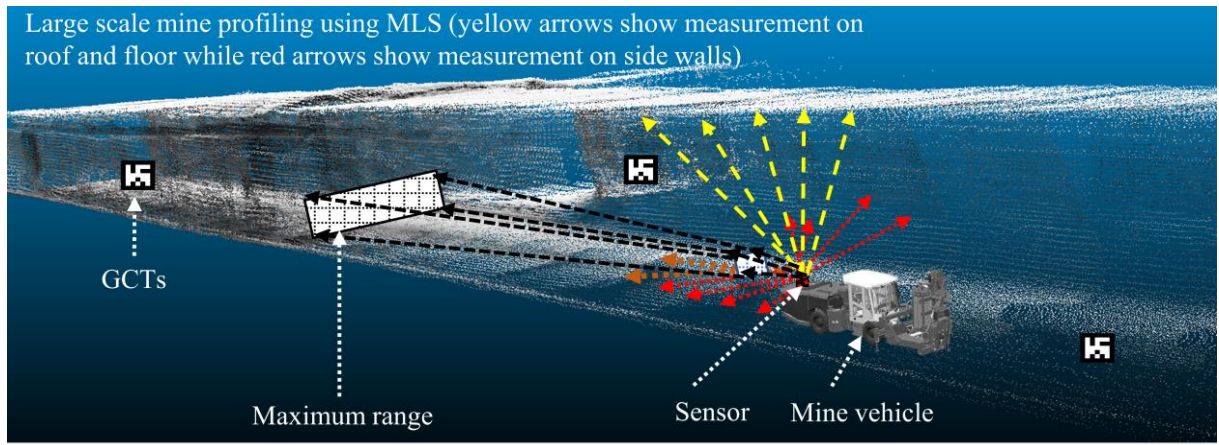

(a) A vehicle-mounted laser scanner system scanning underground roadway with installed GCTs

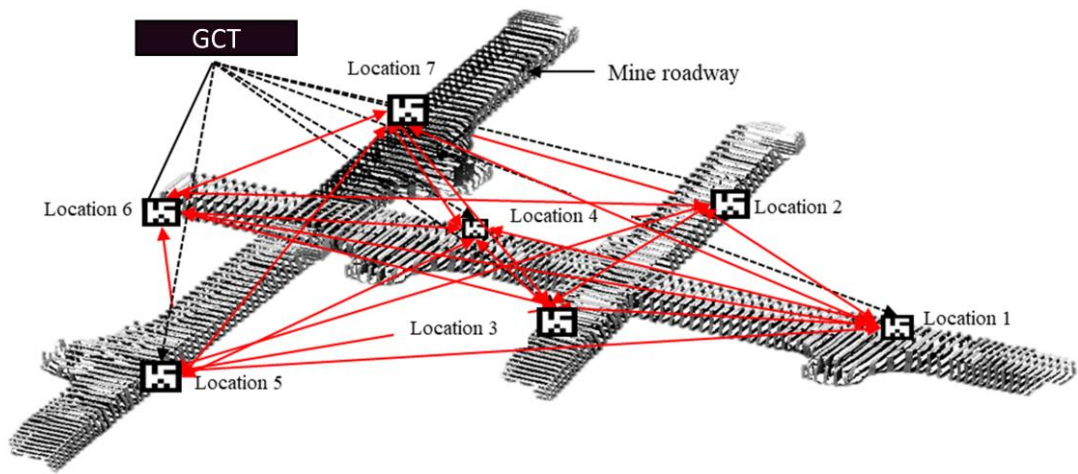

(b) Extraction of roadway clearance between a given network of GCT for comparison in multi-temporal data

**Fig. 9.** Concept of multi-temporal roadway clearance comparison for measuring convergence and floor heave rate (after [12]).

## 5.6  Support structure monitoring

Hazardous ground movements and weak rock strata compromise the structural stability of an underground mine, leading to events such as rockfalls or roof falls. To avoid such events, the load-bearing capacity of weak rock strata is strengthened by using structural supports such as rock or cable bolts and by applying reinforced shotcrete with metal mesh or steel fibres on the walls and roof. Roof falls due to weak structural support may cause fatal injuries, impair the ventilation or disrupt mining operations, leading to expensive downtime [158]. Improper installation, inappropriate design, changes due to mining activities and corrosion of support structures over time are among the enumerable factors leading to structural instabilities, which necessitate routine monitoring of support structures and modelling of the related structural strength of the rock mass [161]. Traditional instrumentation for this purpose, such as a ring gauge, strain gauge, extensometer, ultrasonic bond tester, shear meter and compression pad, enable monitoring at discrete spatial locations, which are not scalable over a larger area. Alternatively, laser scanning provides an attractive option to monitor large mine spaces and help visualise the density, displacement and change in rock bolts over time.

Benton et al. [162] proposed the concept of using a point cloud obtained from structure from motion to monitor corrosion of rock bolts as a direction for future research. The optical information captured through photogrammetry or intensity information captured in laser scans could be used to estimate the extent of corrosion of rock bolts. Such interest in identifying roof bolts from point-cloud scans has emerged only recently. These methods use local point descriptors to capture cylinder-like properties of rock bolts for automated detection. In one instance, Gallwey et al. [149] defined 61 descriptors for each point in the point cloud to capture surrounding information, and then a neural network was trained on the defined descriptors to classify roof bolts (Fig. 10a). The approach resulted in a precision of 93% and recall of 87% on a point cloud captured from TLS. A robust approach for identifying rock bolts in a 3D point cloud captured from MLS was also presented by Singh et al. [133]. Since MLS has low data density, the workflow incorporated intensive pre-processing through iterative point-cloud resampling and moving least-squares projection techniques to obtain adequate point spacing. Then, a support vector machine was trained on multi-scale eigenvalue point descriptors to identify rock bolts, resulting in precision, recall and quality of 86.73%, 89.27% and 78.54%, respectively, on the validation dataset. The study also presented a topological map of roof bolt density (Fig. 10b) and distance (Fig. 10c) for a visual representation of roof bolt distribution within an area. A higher precision (93%) and recall (87%) were then achieved by Gallwey et al. [163], as TLS provided better point density and lower noise.

We can note that using many features does not necessarily help, as bolts and non-bolt points may exhibit similar behaviours under certain descriptor categories (such as omnivariance, verticality) over a given scale. Such descriptors act as noise and often mislead the classification, and generating many descriptors is time-consuming. Accuracies could be improved instead by using selective point descriptors that show high variance for bolt and non-bolt classes. Singh et al. [164] investigated a specific set of descriptors (namely radial surface descriptors), the proportion of variance and fast point feature histogram on an MLS point cloud to capture the cylindrical property of roof bolts, for classification in an underground coal mine. Among the compared machine-learning algorithms, ANN resulted in the highest precision (89.52%), recall (89.52%) and quality (81.03%). Building on point descriptors, Saydam et al. [165] proposed the PointNet deep learning algorithm "CFBolt" to segregate rock bolts. First, the majority of points were removed using proportion of variance local descriptors, then a deep learning algorithm was applied to identify candidate rock bolt points. For a TLS scan, CFBolt resulted in a precision of 98.2%, recall of 98.7% and F1 score of 98.5.

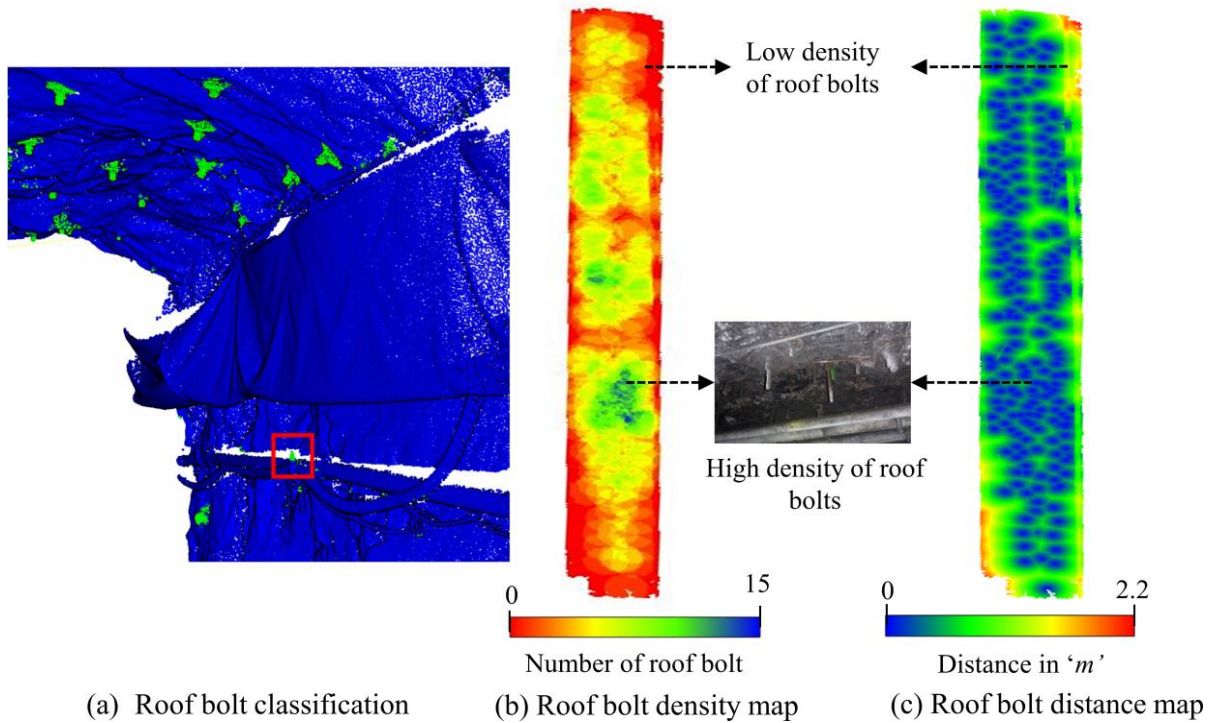

**Fig. 10.** Roof bolt identification in 3D point cloud data (modified from Gallwey et al. [149] and Singh et al. [122]).

Mapping and evaluating the shotcrete thickness is a common monitoring and quality-assurance measure in underground mines for improved structural safety. This thickness is the governing parameter that determines the flexural capacity of underground mines or tunnels [166]. Shotcrete-thickness mapping involves aligning multi-temporal point clouds, which are then compared to evaluate the radial thickness of shotcrete. Fekete et al. [167] used TLS in a drill and blast tunnel to measure the shotcrete condition through multi-temporal scans. The thickness was measured using laser scans collected before and after applying shotcrete. Additionally, water seepage through fractures was detected using intensity data from the scans, as damp shotcrete tends to appear darker in the point cloud (low-intensity values) due to absorption of NIR.

Further research on shotcrete thickness using laser scanning suggested that the data should be collected on the same day as a larger time difference leads to radial mine convergence depending on the rock mass quality, strength, stress, size of the opening and support [168]. A scan comparison that ignores the convergence rate often overestimates the shotcrete thickness, giving a false value for the safety factor. It is possible to account for convergence either through multi-temporal data collection or numerical modelling. Lato and Diederichs [168] presented three-step scanning to account for convergence when calculating the shotcrete thickness. The first and second scans were collected before applying shotcrete and the final scan was collected after. The approach effectively quantified the convergence rate, thus improving the accuracy of shotcrete thickness calculation. The accuracy of the shotcrete thickness determined also depends on the point cloud registration; in this vein, past studies have used local geometric features [169] and artificially installed spherical targets [170] to reduce the registration error when measuring the shotcrete thickness.

*5.7  Stope dilution and analysis of blasting-formed underground cavities*

Open stoping is a high-production, low-cost underground mining method to extract ore. Yet, unplanned stope/ore dilution is a major issue with the open stoping method that affects its efficiency. Dilution leads to the mixing of ore with barren waste, which lowers the mill recovery and effective capacity as the mine ages [171]. Consequently, there is a direct influence on the profitability of a mining operation owing to costs associated with the mucking, haulage crushing, hoisting, milling and treatment of ore, besides production delays. Dilution and recovery of a stope are generally considered a measure of the quality of stope design and the mining performance. Dilution is estimated in terms of the average volume of unplanned material that has sloughed off the stope walls within the unit area. This process is referred to as equivalent linear overbreak or slough [172]. Conversely, it may also encompass material intended to be mined but not recovered, known as underbreak. Major factors influencing unplanned dilution include geology and rock mechanics, stope design, drilling, blasting and human assessment errors [171].

It is critically important to study factors leading to dilution, and quantitatively estimate overbreaks and underbreaks, to assess the stability, overall performance and return on investments. According to a recent industry survey [173], stope assessment primarily relies on a cavity monitoring system (CMS) that uses laser scanning as a base technology and provides a point cloud as an output. Henning and Mitri [174] compiled a comprehensive database of parameters derived from a laser-scanning CMS to investigate the factors causing ore dilution. The study concluded that, in addition to stope dimensions, unplanned dilution is dependent on the stope type as the overbreak varies. In a similar study, Amedjoe and Agyeman [175] studied the overbreak and underbreak for ore dilution by comparing the 3D void model of stope with the designed stope boundaries. The study presented a quantitative estimate of ore losses due to overbreak and underbreak that contributed to an 11% loss upon the completion of mining activities.

Beyond this, a laser scanning system is also crucial to assess the stability of stopes and has been used to verify the dependence of stability on the distribution of joints [176]. Automated mapping of joint features in stopes using a point cloud is complex due to the enclosed (cylindrical) nature of stopes and difficulty in distinguishing natural and artificial planar features. In that context, studies use manual identification of joint planes in a 3D point cloud for stability analysis [132,176]. A CMS is currently the most preferred way to monitor stopes, but one critical disadvantage lies in the presence of blind spots due to static sensing. To resolve blind spot issues, recently, drone-mounted SLAM-based laser scanners, such as Hovermap, were investigated for their potential to generate more comprehensive and accurate 3D data [21]. Such systems are now being used routinely to monitor stope volumes, provide insights into production stocks and recognise sources of instability [21]. When considering the current state-of-the-art technology, laser scanning has been identified as the best option for stope assessment. In the years to come, other assessment methods, such as visual inspection, seismic controls and borehole extensometers, when combined with laser scanning, could support the development of a more robust stope monitoring solution [173].

Large-scale blasting in underground mines regulates the efficiency of rock fragmentation in an area and affects the natural continuity of rocks beyond the blasting perimeter [177]. The blasting efficiency in underground mines is governed by the type and level of stress and the hierarchical block structure of a rock mass. Laser scanning plays an important role in analysing the blasting efficiency, which greatly impacts the geomechanical behaviour of a rock mass and zonal disintegration around underground excavations. In one instance, TLS was used in an iron ore mine to determine the stope configuration, mainly the angular parameter of the roof and cross-sectional shape, which was correlated with the stress state and structure of the surrounding rock mass. The use of TLS allowed for the evaluation of rock mass disintegration, to thereby estimate the blasting efficiency [177]. Similarly, blasting-induced crater characteristics such as displaced rock volumes and breakout shapes were analysed using a TLS survey to evaluate the blast efficiency [178]. In other work, laser scanning was also used to measure rock fragments in underground mines, to assess the blasting efficiency and evaluate the performance of loading equipment and crushing systems [179]. The process involves object-based classification of rock fragments by delineating the boundaries of individual rocks. Then, a histogram of rock size is created to evaluate the blasting. Here, surface characteristics obtained from laser scanning, such as rock fragments' sizes, structure measurements and lithological information, could significantly improve the efficiency of underground mines. For instance, research developing empirical models to correlate surface measurements captured from laser scanners could be used to evaluate the performance of blasting or excavation systems.

## 6  Miscellaneous applications and future directions

Three-dimensional models obtained from a laser scanner have become reliable tools in addressing areas of concern identified by mine safety bodies. Apart from the essential applications discussed above, potential uses of laser scanning are being explored to widen its applicability in underground mining. For instance, accurate cemented paste backfilling in the processing of waste tailings and underground goaf was initially difficult due to the shortage of monitoring devices or techniques that could work in a harsh environment. With the availability of laser scanning, the exact height of cement backfill can now be estimated besides monitoring real-time backfilling progress using accurate volumetric estimation of goaf areas [180]. In another study, Watson and Marshall [181] proposed using 3D point clouds obtained from MLS (uGPS mapper) to calculate friction factors for mine ventilation in a mine in Northern Ontario, Canada. The roughness measure was obtained by calculating the variance of points in the cross-sectional and longitudinal directions, the mean of which was considered the equivalent roughness ($\varepsilon_t$) when calculating the mine friction to airflow. The results were compared to VelociCalc Multi-Function Ventilation Meter model 9565, exhibiting comparatively higher accuracy than conventional methods. The versatility of laser scanning was also demonstrated by integrating a point cloud with other sensor measurements such as gas concentrations, wind speed, temperature, humidity, pressure, etc. [182]. The point cloud was coloured with respect to measurements from sensors, to provide an interactive visual representation of varying conditions across the mine site. Moreover, a fly-through video of an underground mine was demonstrated to show the potential for remote management of logistics and hazards.

Though not yet researched extensively, investigations of various subsidence-related factors in an underground coal mine are of critical importance and could be facilitated by laser scanning. The main subsidence-related factors are surface basin parameters and geomorphic features [183]. Surface-basin parameters mainly include horizontal and vertical displacement, tilting, curvature, critical/subcritical/supercritical panel widths, the angle of draw, width-to-depth ratio, tension cracks, turf rolls and pushouts. Similarly, geomorphic features encompass landslides, heaves, rockfalls, sinkholes, troughs, cracks/fissures and linear sinks. Most of the above parameters can be determined using laser scanning, which could lead potential hazards to be identified. Beyond this, from a management perspective, it is challenging to track logistics due to the large underground spaces, poor lighting and GNSS-related constraints [8,18,22]. Laser scanning could play a pivotal role in solving such issues through accurate object detection and semantic segmentation of mine roadways.

Since there is a lack of training data that could facilitate such applications using deep learning, targeted algorithms must be developed to track objects of importance in 3D point clouds, besides creating an object repository. Applications such as automated deformation monitoring, autonomous navigation, object detection, search and rescue, volumetric computation, roadway convergence monitoring, structure mapping and fragmentation analysis would greatly benefit from such a data repository. Then, as a separate avenue of interest, recently, smartphone LiDAR sensors have become more popular due to their simplicity of use and low cost. However, the reliability of smartphone-based LiDAR has not yet been demonstrated for geological and geotechnical applications in underground mines. Yet, the technology is rapidly evolving, and once proven, smartphone-based laser scanning could significantly reduce underground monitoring costs. Furthermore, with advancements in graphical processing units (GPUs) and low-power single-board computers, it is now possible to develop edge computing workflows that could become essential for onsite real-time assessments.

## 7 Conclusion

The underground mining industry is increasingly adopting TLS and MLS as routine monitoring devices owing to the reductions they bring in sensor costs, as they represent robust solutions in technically challenging conditions and thanks to their large-scale data-capture capabilities. In this study, the current state of laser-scanning solutions in mining applications was critically evaluated, and new insights were provided into progressive developments that are moving the field towards automation and digital twin. Moreover, the study highlighted the key needs and requirements that are pushing stakeholders to widen the applicability of laser scanning in underground mines. The review of scientifically peer-reviewed studies indicated that some areas, such as rock mass characterisation, convergence monitoring, shotcrete thickness mapping and change detection, have been extensively researched. In contrast, other areas, such as logistic management, lithological classification and investigation of surface basin parameters, remain less explored. Nevertheless, laser scanning shows huge potential in solving some of the most critical issues from production and hazard-management perspectives.

The accuracy requirements and scale of application govern the selection of TLS or MLS. Accordingly, applications such as convergence monitoring, measuring the floor heave rate or tracking millimetre-level deformations that threaten rockfalls prefer TLS. MLS, meanwhile, in its current state, may not represent a replacement for TLS in terms of accuracy. Nevertheless, given recent advancements in SLAM algorithms, coupled with the development of new positioning sensors for GNSS-denied environments, its accuracy is expected to improve. That will be advantageous as the simplicity of data collection, cost-effectiveness and mobility offered by MLS outweigh the applicability of TLS for large-scale routine monitoring of underground mines. Another vital aspect is the processing capability in underground mines. At present, due to the huge amount of data captured from a laser scanner, processing is performed off-site with high-end computing equipment. However, by optimising processing algorithms and implementing GPU for some of the workflow, it is increasingly becoming possible to process data on-site if a quick analysis of the site is required. Currently, in some cases, particularly in coal mines, the applicability may be restricted by intrinsic safety constraints and fire-related hazards. Developments in safety and fire-proof enclosures aim to tackle such issues, but are yet to be tested in sensitive underground mines for their accuracy and robustness in data collection. At the same time, smartphone-based low-power LiDAR is gaining popularity, and it will be interesting to compare its performance with conventional TLS and MLS systems in the years to come.

The findings from this review suggest there is scope to improve processing algorithms to increase their efficiency and accuracy for already well-researched areas such as change detection and rock mass characterisation. Furthermore, processing algorithms could be fully automated, where parameters are adaptively computed based on the application and point-cloud resolution, to reduce human involvement in parameter selection. From a future work perspective, we recommend that a comprehensive data repository be created of underground objects that could play a pivotal role in semantic segmentation, localisation, logistics management, search and rescue, support structure management and deformation monitoring. Additional data modalities could be included in the data repository that would aid in multi-sensor integration and monitoring of mine sites. With the mining industry moving towards automation, it is our impression that laser scanning will become critical to facilitate remote and autonomous mine management while reducing the human footprint in operational areas.

## References

[1]   Mukupa W, Roberts GW, Hancock CM, Al-Manasir K. A review of the use of terrestrial laser scanning application for change detection and deformation monitoring of structures. Surv Rev 2017;49:99–116. https://doi.org/10.1080/00396265.2015.1133039.

[2]   Sammartano G, Spanò A. Point clouds by SLAM-based mobile mapping systems: accuracy and geometric content validation in multisensor survey and stand-alone acquisition. Appl Geomatics 2018:1–23. https://doi.org/10.1007/s12518-018-0221-7.

[3]   Raval S, Banerjee BP, Kumar Singh S, Canbulat I. A Preliminary Investigation of Mobile Mapping Technology for Underground Mining. In: Proceedings of the IEEE International Geoscience and Remote Sensing Symposium. Yokohama: Institute of Electrical and Electronics Engineers; 2019.p.6071–4. https://doi.org/10.1109/IGARSS.2019.8898518.

[4]   Wong U, Morris A, Lea C, Lee J, Whittaker C, Garney B, Whittaker R. Comparative evaluation of range sensing technologies for underground void modeling. In: Proceedings of the IEEE/RSJ International Conference on Intelligent Robots and Systems. San Francisco: Institute of Electrical and Electronics Engineers; 2011.p.3816–23. https://doi.org/10.1109/IROS.2011.6048626.

[5]   Lai P, Samson C. Applications of mesh parameterization and deformation for unwrapping 3D images of rock tunnels. Tunn Undergr Sp Technol 2016;58:109–19.

https://doi.org/10.1016/j.tust.2016.04.009.

[6]  Kukutsch R, Kajzar V, Konicek P, Waclawik P, Ptacek J. Possibility of convergence measurement of gates in coal mining using terrestrial 3D laser scanner. J Sustain Min 2015;14:30–7. https://doi.org/10.1016/j.jsm.2015.08.005.

[7]  Dewez TJB, Yart S, Thuon Y, Pannet P, Plat E. Towards cavity-collapse hazard maps with Zeb-Revo handheld laser scanner point clouds. Photogramm Rec 2017;32:354–76. https://doi.org/10.1111/phor.12223.

[8]  Ellmann A, Kütimets K, Varbla S, Väli E, Kanter S. Advancements in underground mine surveys by using SLAM-enabled handheld laser scanners. Surv Rev 2021;0:1–12. https://doi.org/10.1080/00396265.2021.1944545.

[9]  Van Der Merwe JW, Andersen DC. Applications and benefits of 3D laser scanning for the mining industry. J South African Inst Min Metall 2013;113:213–9.

[10] Pfeifer N, Dorninger P, Haring A, Fan H. Investigating terrestrial laser scanning intensity data: quality and functional relations. In: Proceedings of the 8th Conference on Optical 3-D Measurement Techniques. Zurich: ETH Zurich; 2007.p.328–37.

[11] Roca-Pardiñas J, Argüelles-Fraga R, de Asís López F, Ordóñez C. Analysis of the influence of range and angle of incidence of terrestrial laser scanning measurements on tunnel inspection. Tunn Undergr Sp Technol 2014;43:133–9. https://doi.org/10.1016/j.tust.2014.04.011.

[12] Singh SK, Raval S, Banerjee BP. Automated structural discontinuity mapping in a rock face occluded by vegetation using mobile laser scanning. Eng Geol 2021;285:106040. https://doi.org/10.1016/j.enggeo.2021.106040.

[13] Prikhodko IP, Bearss B, Merritt C, Bergeron J, Blackmer C. Towards self-navigating cars using MEMS IMU: Challenges and opportunities. In: Proceedings of the IEEE International Symposium on Inertial Sensors and Systems (INERTIAL). Lake Como: Institute of Electrical and Electronics Engineers; 2018.p.1–4. https://doi.org/10.1109/ISISS.2018.8358141.

[14] Zhang Q, Niu X, Shi C. Impact Assessment of Various IMU Error Sources on the Relative Accuracy of the GNSS/INS Systems. IEEE Sens J 2020;20:5026–38. https://doi.org/10.1109/JSEN.2020.2966379.

[15] Unsal D, Demirbas K. Estimation of deterministic and stochastic IMU error parameters. In: Proceedings of the IEEE/ION Position, Location and Navigation Symposium. Myrtle Beach: Institute of Electrical and Electronics Engineers; 2012.p. 862–8. https://doi.org/10.1109/PLANS.2012.6236828.

[16] Kolecki J, Prochaska M, Kurczynski Z, Piatek P, Baranowski J. Developing the stabilized mapping system for the gyrocopter - Report from the first tests. Int Arch Photogramm Remote Sens Spat Inf Sci - ISPRS Arch 2016;2016-Janua:31–6. https://doi.org/10.5194/isprsarchives-XLI-B1-31-2016.

[17] Voisin S, Foufou S, Truchetet F, Page DL, Abidi MA. Study of ambient light influence for three-dimensional scanners based on structured light. Opt Eng 2007;46:1–3. https://doi.org/10.1117/1.2717126.

[18] Singh SK, Banerjee BP, Raval S. Three-dimensional unique-identifier-based automated

georeferencing and coregistration of point clouds in underground mines. Remote Sens 2021;13. https://doi.org/10.3390/rs13163145.

[19] Soudarissanane S, Lindenbergh R, Menenti M, Teunissen P. Scanning geometry: Influencing factor on the quality of terrestrial laser scanning points. ISPRS J Photogramm Remote Sens 2011;66:389–99. https://doi.org/10.1016/J.ISPRSJPRS.2011.01.005.

[20] Dissanayake G, Huang S, Wang Z, Ranasinghe R. A review of recent developments in Simultaneous Localization and Mapping. In: Proceedings of the 6th International Conference on Industrial and Information Systems. Kandy: Institute of Electrical and Electronics Engineers; 2011.p. 477–82. https://doi.org/10.1109/ICIINFS.2011.6038117.

[21] Jones E, Sofonia J, Canales C, Hrabar S, Kendoul F. Applications for the Hovermap autonomous drone system in underground mining operations. J South African Inst Min Metall 2020;120:49–56. https://doi.org/10.17159/2411-9717/862/2020.

[22] Chen S, Walske ML, Davies IJ. Rapid mapping and analysing rock mass discontinuities with 3D terrestrial laser scanning in the underground excavation. Int J Rock Mech Min Sci 2018;110:28–35. https://doi.org/10.1016/j.ijrmms.2018.07.012.

[23] Errington AFCC, Daku BLFF, Prugger A. Closure Monitoring in Potash Mines using LiDAR. In: Proceedings of the 36th Annual Conference on IEEE Industrial Electronics Society. Glendale: Institute of Electrical and Electronics Engineers; 2010.p.2823–7. https://doi.org/10.1109/IECON.2010.5675071.

[24] Biber P, Strasser W. The normal distributions transform: a new approach to laser scan matching. IEEE Int. Conf. Intell. Robot. Syst., vol. 3, 2003, p. 2743–8. https://doi.org/10.1109/iros.2003.1249285.

[25] Magnusson M, Andreasson H, Nuchter A. Automatic Appearance-Based Loop Detection from Three-Dimensional Laser Data Using the Normal Distributions Transform. J F Robot 2009;26:892–914. https://doi.org/10.1002/rob.20314.

[26] Bailey T, Durrant-Whyte H. Simultaneous localization and mapping (SLAM): Part I. IEEE Robot Autom Mag 2006;13:108–17. https://doi.org/10.1109/MRA.2006.1678144.

[27] Cadena C, Carlone L, Carrillo H, Latif Y, Scaramuzza D, Neira J, Reid I, Leonard JJ. Past, present, and future of simultaneous localization and mapping: Toward the robust-perception age. IEEE Trans Robot 2016;32:1309–32. https://doi.org/10.1109/TRO.2016.2624754.

[28] Arshad S, Kim GW. Role of deep learning in loop closure detection for visual and lidar slam: A survey. Sensors (Switzerland) 2021;21:1–17. https://doi.org/10.3390/s21041243.

[29] Kalman RE, R. Kalman. A New Approach to Linear Filtering and Prediction Problems. Trans ASME-Journal Basic Eng 1960;82:35–45. https://doi.org/DOI:10.1115/1.3662552.

[30] Bharani Chandra KP, Gu DW, Postlethwaite I. Cubature Kalman filter based Localization and Mapping. In: Proceedings of the 18th World Congress of the International Federation of Automatic Control. Milano: International Federation of

Automatic Control; 2011.p. 2121–5. https://doi.org/10.3182/20110828-6-IT-1002.03104.

[31] Angelo M. Sabatini. Quaternion-based extended Kalman filter for determining orientation by inertial and magnetic sensing. IEEE Trans Biomed Eng 2006;53:1346–56.

[32] Eustice RM, Singh H, Leonard JJ. Exactly sparse delayed-state filters. In: Proceedings of the IEEE International Conference on Robotics and Automation. Barcelona: Institute of Electrical and Electronics Engineers; 2005.p.2417–24. https://doi.org/10.1109/ROBOT.2005.1570475.

[33] Thrun S. Probabilistic robotics. Commun ACM 2002;45:1999–2000. https://doi.org/10.1145/504729.504754.

[34] Montemerlo M, Thrun S. FastSLAM 1.0. FastSLAM A Scalable Method Simultaneous Localization Mapp Probl Robot 2007:27–62.

[35] Li D, Zhou B, Wang Z, Yang S, Liu P. The Identification and Compensation of Static Drift Induced by External Disturbances for LiDAR SLAM. IEEE Access 2021;9:58102–15. https://doi.org/10.1109/ACCESS.2021.3072935.

[36] Stachniss C, Kretzschmar H. Pose graph compression for laser-based SLAM. Springer Tracts Adv. Robot., 2017. https://doi.org/10.1007/978-3-319-29363-9_16.

[37] Pinto AM, Moreira AP, Costa PG. A Localization Method Based on Map-Matching and Particle Swarm Optimization. J Intell Robot Syst Theory Appl 2013;77:313–26. https://doi.org/10.1007/s10846-013-0009-2.

[38] Zhang J, Singh S. LOAM: Lidar Odometry and Mapping in Real-time. Robot. Sci. Syst., vol. 2, 2014.

[39] Ćwian K, Nowicki MR, Wietrzykowski J, Skrzypczyński P. Large-scale lidar slam with factor graph optimization on high-level geometric features. Sensors 2021;21. https://doi.org/10.3390/s21103445.

[40] Latif Y, Cadena C, Neira J. Robust loop closing over time for pose graph SLAM. Int J Rob Res 2013;32:1611–26. https://doi.org/10.1177/0278364913498910.

[41] Shin DW, Ho YS. Loop closure detection in simultaneous localization and mapping using learning based local patch descriptor. In: Proceedings of the International Symposium on Electronic Imaging: Autonomous Vehicles and Machines. Burlingame: Society for Imaging Science and Technology; 2018.p.284–1. https://doi.org/10.2352/ISSN.2470-1173.2018.17.AVM-284.

[42] Chen C, Wang B, Lu CX, Trigoni N, Markham A. A Survey on Deep Learning for Localization and Mapping: Towards the Age of Spatial Machine Intelligence 2020.

[43] Ren Z, Wang L, Bi L. Robust GICP-Based 3D LiDAR SLAM for Underground Mining Environment. Sensors 2019;19. https://doi.org/10.3390/s19132915.

[44] Droeschel D, Behnke S. Efficient Continuous-Time SLAM for 3D Lidar-Based Online Mapping. In: Proceedings of the IEEE International Conference on Robotics and Automation. Brisbane: Institute of Electrical and Electronics Engineers; 2018.p.5000–7. https://doi.org/10.1109/icra.2018.8461000.

[45] Mur-Artal R, Montiel JMM, Tardos JD. ORB-SLAM: A Versatile and Accurate Monocular SLAM System. IEEE Trans Robot 2015;31:1147–63. https://doi.org/10.1109/TRO.2015.2463671.

[46] Chu T, Guo N, Backén S, Akos D. Monocular camera/IMU/GNSS integration for ground vehicle navigation in challenging GNSS environments. Sensors 2012;12:3162–85. https://doi.org/10.3390/s120303162.

[47] López E, García S, Barea R, Bergasa LM, Molinos EJ, Arroyo R, Romera E, Pardo S. A multi-sensorial simultaneous localization and mapping (SLAM) system for low-cost micro aerial vehicles in GPS-denied environments. Sensors (Switzerland) 2017;17. https://doi.org/10.3390/s17040802.

[48] Leung K, Lühr D, Houshiar H, Inostroza F, Borrmann D, Adams M, Nüchter A, Ruiz Del Solar J, Ruiz J. Chilean underground mine dataset. Int J Rob Res 2017;36:16–23. https://doi.org/10.1177/0278364916679497.

[49] Mendes E, Koch P, Lacroix S. ICP-based pose-graph SLAM. In: Proceedings of the IEEE International Symposium on Safety, Security, and Rescue Robotics. Lausanne: Institute of Electrical and Electronics Engineers; 2016.p.195–200. https://doi.org/10.1109/SSRR.2016.7784298.

[50] Hsu YW, Huang SS, Perng JW. Application of multisensor fusion to develop a personal location and 3D mapping system. Optik (Stuttg) 2018;172:328–39. https://doi.org/10.1016/j.ijleo.2018.07.029.

[51] Kaasalainen S, Ruotsalainen L, Kirkko-Jaakkola M, Nevalainen O, Hakala T. Towards multispectral, multi-sensor indoor positioning and target identification. Electron Lett 2017;53:1008–11. https://doi.org/10.1049/el.2017.1473.

[52] Ghosh D, Samanta B, Chakravarty D. Multi sensor data fusion for 6D pose estimation and 3D underground mine mapping using autonomous mobile robot. Int J Image Data Fusion 2017;8:173–87. https://doi.org/10.1080/19479832.2016.1226966.

[53] Jacobson A, Zeng F, Smith D, Boswell N, Peynot T, Milford M. What localizes beneath: A metric multisensor localization and mapping system for autonomous underground mining vehicles. J F Robot 2021;38:5–27. https://doi.org/10.1002/rob.21978.

[54] Neumann T, Ferrein A, Kallweit S, Scholl I. Towards a Mobile Mapping Robot for Underground Mines. In: Proceedings of the 2014 PRASA, RobMech and AfLaT International Joint Symposium. Cape Town: Pattern Recognition Association of South Africa; 2014.p. 27–8.

[55] Zlot R, Bosse M. Efficient large-scale 3D mobile mapping and surface reconstruction of an underground mine. F. Serv. Robot., Springer; 2014, p. 479–93.

[56] Papachristos C, Khattak S, Mascarich F, Alexis K. Autonomous Navigation and Mapping in Underground Mines Using Aerial Robots. In: Proceedings of the IEEE Aerospace Conference. Big Sky: Institute of Electrical and Electronics Engineers; 2019.p.1–8. https://doi.org/10.1109/AERO.2019.8741532.

[57] Mansouri SS, Kanellakis C, Kominiak D, Nikolakopoulos G. Deploying MAVs for autonomous navigation in dark underground mine environments. Rob Auton Syst 2020;126:103472. https://doi.org/10.1016/j.robot.2020.103472.

[58] Rubio-Sierra C, Domínguez D, Gonzalo J, Escapa A. Path planner for autonomous exploration of underground mines by aerial vehicles. Sensors (Switzerland) 2020;20:1–27. https://doi.org/10.3390/s20154259.

[59] Dang T, Mascarich F, Khattak S, Nguyen H, Nguyen H, Hirsh S, Reinhart R, Papachristos C, Alexis K. Autonomous Search for Underground Mine Rescue Using Aerial Robots. In: Proceedings of the IEEE Aerospace Conferences. Big Sky: Institute of Electrical and Electronics Engineers; 2020.p. 1–8. https://doi.org/10.1109/AERO47225.2020.9172804.

[60] Kim H, Choi Y. Location estimation of autonomous driving robot and 3D tunnel mapping in underground mines using pattern matched LiDAR sequential images. Int J Min Sci Technol 2021;31:779–88. https://doi.org/https://doi.org/10.1016/j.ijmst.2021.07.007.

[61] Li M, Zhu H, You S, Wang L, Tang C. Efficient Laser-Based 3D SLAM for Coal Mine Rescue Robots. IEEE Access 2019;7:14124–38. https://doi.org/10.1109/ACCESS.2018.2889304.

[62] Eyre M, Weltherelt A, Coggan J, Wetherelt A, Coggan J. Evaluation of automated underground mapping solutions for mining and civil engineering applications. J Appl Remote Sens 2016;10:025013. https://doi.org/10.1117/1.JRS.10.

[63] Dunn M, Reid P, Malos J. Development of a protective enclosure for remote sensing applications-Case study: Laser scanning in underground Coal Mines. Resources 2020;9. https://doi.org/10.3390/RESOURCES9050056.

[64] Pejić M, Božič B, Abolmasov B, Gospavić Z. Design and optimisation of laser scanning for tunnels geometry inspection. Tunn Undergr Sp Technol 2013;37:199–206. https://doi.org/10.1016/j.tust.2013.04.004.

[65] Xing Z, Zhao S, Guo W, Guo X, Wang Y. Processing Laser Point Cloud in Fully Mechanized Mining Face Based on DGCNN. ISPRS Int J Geo-Information 2021;10:482. https://doi.org/10.3390/ijgi10070482.

[66] Besl PJ, McKay ND. A method for registration of 3D shapes. IEEE Trans Pattern Anal Mach Intell 1992;14:239–56. https://doi.org/10.1109/34.121791

[67] Dong Z, Liang F, Yang B, Xu Y, Zang Y, Li J, Wang Y, Dai W, Fan H, Hyyppäb J, Stilla U. Registration of large-scale terrestrial laser scanner point clouds: A review and benchmark. ISPRS J Photogramm Remote Sens 2020;163:327–42. https://doi.org/10.1016/j.isprsjprs.2020.03.013.

[68] Eo YD, Pyeon MW, Kim SW, Kim JR, Han DY. Coregistration of terrestrial lidar points by adaptive scale-invariant feature transformation with constrained geometry. Autom Constr 2012;25:49–58. https://doi.org/10.1016/j.autcon.2012.04.011.

[69] Rusu RB, Blodow N, Beetz M. Fast Point Feature Histograms (FPFH) for 3D registration. In: Proceedings of the IEEE International Conference on Robotics and Automation. Kobe: Institute of Electrical and Electronics Engineers; 2009.p.3212–7. https://doi.org/10.1109/robot.2009.5152473.

[70] Yang J, Cao Z, Zhang Q. A fast and robust local descriptor for 3D point cloud registration. Inf Sci (Ny) 2016;346–347:163–79.

https://doi.org/10.1016/j.ins.2016.01.095.

[71] Dong Z, Yang B, Liang F, Huang R, Scherer S. Hierarchical registration of unordered TLS point clouds based on binary shape context descriptor. ISPRS J Photogramm Remote Sens 2018;144:61–79. https://doi.org/10.1016/j.isprsjprs.2018.06.018.

[72] Zhang Z, Dai Y, Sun J. Deep learning based point cloud registration: an overview. Virtual Real Intell Hardw 2020;2:222–46. https://doi.org/10.1016/j.vrih.2020.05.002.

[73] Lu W, Wan G, Zhou Y, Fu X, Yuan P, Song S. DeepVCP: An end-to-end deep neural network for point cloud registration. In: Proceedings of the IEEE/CVF International Conference on Computer Vision. Seoul: Institute of Electrical and Electronics Engineers; 2019.p.12–21. https://doi.org/10.1109/ICCV.2019.00010.

[74] Alarifi A, Al-Salman A, Alsaleh M, Alnafessah A, Al-Hadhrami S, Al-Ammar MA, Al-Khalifa HS. Ultra wideband indoor positioning technologies: Analysis and recent advances. Sensors (Switzerland) 2016;16:1–36. https://doi.org/10.3390/s16050707.

[75] Wang Z, Wu LX, Li HY. Key technology of mine underground mobile positioning based on LiDAR and coded sequence pattern. Trans Nonferrous Met Soc China (English Ed 2011;21:s570–6. https://doi.org/10.1016/S1003-6326(12)61642-2.

[76] Simela JV, Marshall JA, Daneshmend LK. Automated laser scanner 2D positioning and orienting by method of triangulateration for underground mine surveying. In: Proceedings of the International Symposium on Automation and Robotics in Construction. Montreal: The International Association for Automation and Robotics in Construction; 2013.p.708–17. https://doi.org/10.22260/isarc2013/0078.

[77] Pesci A, Teza G. Terrestrial laser scanner and retro-reflective targets: An experiment for anomalous effects investigation. Int J Remote Sens 2008;29:5749–65. https://doi.org/10.1080/01431160802108489.

[78] Hlophe K, Du Plessis F. Implementation of an autonomous underground localization system. In: Proceedings of the 6th Robotics and Mechatronics Conference (RobMech). Durban: Institute of Electrical and Electronics Engineers; 2013.p.87–92. https://doi.org/10.1109/RoboMech.2013.6685497.

[79] Lavigne NJ, Marshall JA, Lavigne, N. James JAM, Lavigne NJ, Marshall JA. A landmark-bounded method for large-scale underground mine mapping. J F Robot 2012;29:861–79. https://doi.org/10.1002/rob.21415.

[80] Martinelli F. Simultaneous Localization and Mapping Using the Phase of Passive UHF-RFID Signals. J Intell Robot Syst Theory Appl 2019;94:711–25. https://doi.org/10.1007/s10846-018-0903-8.

[81] Motroni A, Buffi A, Nepa P. A survey on Indoor Vehicle Localization through RFID Technology. IEEE Access 2021. https://doi.org/10.1109/ACCESS.2021.3052316.

[82] Jung J, Choi Y. Analysis of tag recognition ranges and rates according to reader transmission power levels when tracking machines by RFID in underground mines: an indoor experiment. Geosystem Eng 2017;20:81–7. https://doi.org/10.1080/12269328.2016.1224985.

[83] Farahneh H, Hussain F, Fernando X. A New Alarming System for an Underground Mining Environment Using Visible Light Communications. In: Proceedings of the IEEE

Canada International Humanitarian Technology Conference. Canada: Institute of Electrical and Electronics Engineers; 2017.p.213–7. https://doi.org/10.1109/IHTC.2017.8058191.

[84] Yoshino M, Haruyama S, Nakagawa M. High-accuracy positioning system using visible LED lights and image sensor. In: Proceedings of the IEEE Radio and Wireless Symposium. Orlando: Institute of Electrical and Electronics Engineers; 2008.p.439–42. https://doi.org/10.1109/RWS.2008.4463523.

[85] Muduli L, Mishra DP, Jana PK. Application of wireless sensor network for environmental monitoring in underground coal mines: A systematic review. J Netw Comput Appl 2018;106:48–67. https://doi.org/10.1016/j.jnca.2017.12.022.

[86] Ahmed S, Gagnon, JD, Makhdoom M, Naeem R, Wang J. New methods and equipment for three-dimensional laser scanning, mapping and profiling underground mine cavities. In: Proceedings of the First International Conference on Underground Mining Technology. Sudbury: Australian Centre for Geomechanics; 2017.p. 467–73.

[87] Schaer P, Vallet J. Trajectory adjustment of mobile laser scan data in GPS denied environments. Int Arch Photogramm Remote Sens Spat Inf Sci 2016;40:61–4. https://doi.org/10.5194/isprs-archives-xl-3-w4-61-2016.

[88] Zhang H, Zhang C, Yang W, Chen CY. Localization and navigation using QR code for mobile robot in indoor environment. In: Proceedings of the IEEE International Conference on Robotics and Biomimetics. Zhuhai: Institute of Electrical and Electronics Engineers; 2015.p.2501–6. https://doi.org/10.1109/ROBIO.2015.7419715.

[89] Yang C, Liu L, Luo W, Meng Y, Su W. Identification of barcode beacon and its application in underground mining. ICACTE 2010 - 2010 3rd Int. Conf. Adv. Comput. Theory Eng. Proc., vol. 1, 2010. https://doi.org/10.1109/ICACTE.2010.5579047.

[90] Yang C, Liu L, Luo W, Meng Y, Su W. Identification of barcode beacon and its application in underground mining. In: Proceedings of the 3rd International Conference on Advanced Computer Theory and Engineering. Chengdu: Institute of Electrical and Electronics Engineers; 2010.p. 128–32. https://doi.org/10.1109/ICACTE.2010.5579047.

[91] Lin G, Chen X. A robot indoor position and orientation method based on 2D barcode landmark. J Comput 2011;6:1191–7. https://doi.org/10.4304/jcp.6.6.1191-1197.

[92] Shi G, Tang J, Guan Y, Cheng X. Target selection and development in 3D laser scanning based on sampling interval. In: Proceedings of the 2nd International Conference on Information Science and Engineering. Hangzhou: Institute of Electrical and Electronics Engineers; 2010.p.4110–2. https://doi.org/10.1109/ICISE.2010.5689334.

[93] Wang Y, Shi H, Zhang Y, Zhang D. Automatic registration of laser point cloud using precisely located sphere targets. J Appl Remote Sens 2014;8:83588. https://doi.org/10.1117/1.jrs.8.083588.

[94] Zhang M. Accurate sphere marker-based registration system of 3D point cloud data in applications of shipbuilding blocks. J Ind Intell Inf 2015;3:318–23. https://doi.org/10.12720/jiii.3.4.318-323.

[95] Nocerino E, Menna F, Remondino F, Toschi I, Rodríguez-Gonzálvez P. Investigation

of indoor and outdoor performance of two portable mobile mapping systems. In: Proceedings of the Videometrics, Range Imaging and Applications XIV. Munich: SPIE digital library; 2017.p. 103320I. https://doi.org/10.1117/12.2270761.

[96] Yang S, Liu S, Zhang N, Li G, Zhang J. A fully automatic-image-based approach to quantifying the geological strength index of underground rock mass. Int J Rock Mech Min Sci 2021;140:104585. https://doi.org/10.1016/j.ijrmms.2020.104585.

[97] Humair F, Abellan A, Carrea D, Matasci B, Epard JL, Jaboyedoff M. Geological layers detection and characterisation using high resolution 3D point clouds: Example of a box-fold in the Swiss Jura Mountains. Eur J Remote Sens 2015;48:541–68. https://doi.org/10.5721/EuJRS20154831.

[98] Živec T, Anžur A, Verbovšek T. Determination of rock type and moisture content in flysch using TLS intensity in the Elerji quarry (south-west Slovenia). Bull Eng Geol Environ 2019;78:1631–43. https://doi.org/10.1007/s10064-018-1245-2.

[99] Walton G, Mills G, Fotopoulos G, Radovanovic R, Stancliffe RPW. An approach for automated lithological classification of point clouds. Geosphere 2016;12:1833–41. https://doi.org/10.1130/GES01326.1.

[100] Penasa L, Franceschi M, Preto N, Teza G, Polito V. Integration of intensity textures and local geometry descriptors from Terrestrial Laser Scanning to map chert in outcrops. ISPRS J Photogramm Remote Sens 2014;93:88–97. https://doi.org/10.1016/j.isprsjprs.2014.04.003.

[101] Okhrimenko M, Coburn C, Hopkinson C. Multi-spectral lidar: Radiometric calibration, canopy spectral reflectance, and vegetation vertical SVI profiles. Remote Sens 2019;11. https://doi.org/10.3390/rs11131556.

[102] Morsy S, Shaker A, El-Rabbany A. Multispectral lidar data for land cover classification of urban areas. Sensors (Switzerland) 2017;17. https://doi.org/10.3390/s17050958.

[103] Kong D, Wu F, Saroglou C. Automatic identification and characterization of discontinuities in rock masses from 3D point clouds. Eng Geol 2020;265:105442. https://doi.org/10.1016/j.enggeo.2019.105442.

[104] Thiele ST, Grose L, Samsu A, Micklethwaite S, Vollgger SA, Cruden AR. Rapid, semi-automatic fracture and contact mapping for point clouds, images and geophysical data. Solid Earth 2017;8:1241–53. https://doi.org/10.5194/se-8-1241-2017.

[105] Ferrero AM, Forlani G, Roncella R, Voyat HI. Advanced geostructural survey methods applied to rock mass characterization. Rock Mech Rock Eng 2009;42:631–65. https://doi.org/10.1007/s00603-008-0010-4.

[106] Gigli G, Casagli N. Semi-automatic extraction of rock mass structural data from high resolution LIDAR point clouds. Int J Rock Mech Min Sci 2011;48:187–98. https://doi.org/10.1016/j.ijrmms.2010.11.009.

[107] Dewez TJB, Girardeau-Montaut D, Allanic C, Rohmer J. Facets : A cloudcompare plugin to extract geological planes from unstructured 3d point clouds. Int Arch Photogramm Remote Sens Spat Inf Sci 2016;41:799–804. https://doi.org/10.5194/isprsarchives-XLI-B5-799-2016.

[108] Vöge M, Lato MJ, Diederichs MS. Automated rockmass discontinuity mapping from 3-

dimensional surface data. Eng Geol 2013;164:155–62. https://doi.org/10.1016/j.enggeo.2013.07.008.

[109] Riquelme AJ, Abellán A, Tomás R, Jaboyedoff M. A new approach for semi-automatic rock mass joints recognition from 3D point clouds. Comput Geosci 2014;68:38–52. https://doi.org/10.1016/j.cageo.2014.03.014.

[110] Guo J, Liu S, Zhang P, Wu L, Zhou W, Yu Y. Towards semi-automatic rock mass discontinuity orientation and set analysis from 3D point clouds. Comput Geosci 2017;103:164–72. https://doi.org/10.1016/j.cageo.2017.03.017.

[111] Zhang P, Du K, Tannant DD, Zhu H, Zheng W. Automated method for extracting and analysing the rock discontinuities from point clouds based on digital surface model of rock mass. Eng Geol 2018;239:109–18. https://doi.org/10.1016/j.enggeo.2018.03.020.

[112] Kong D, Wu F, Saroglou C, Sha P, Li B. In-situ block characterization of jointed rock exposures based on a 3D point cloud model. Remote Sens 2021;13. https://doi.org/10.3390/rs13132540.

[113] Li X, Chen Z, Chen J, Zhu H. Automatic characterization of rock mass discontinuities using 3D point clouds. Eng Geol 2019;259:105131. https://doi.org/10.1016/j.enggeo.2019.05.008.

[114] Wang X, Zou LJ, Shen XH, Ren YP, Qin Y. A region-growing approach for automatic outcrop fracture extraction from a three-dimensional point cloud. Comput Geosci 2017;99:100–6. https://doi.org/10.1016/j.cageo.2016.11.002.

[115] Petrovic S. A comparison between the silhouette index and the davies-bouldin index in labelling ids clusters. In: Proceedings of the 11th Nordic workshop of secure IT systems. Linköping: NordSec; 2006.p. 53–64.

[116] Chen J, Zhu H, Li X. Automatic extraction of discontinuity orientation from rock mass surface 3D point cloud. Comput Geosci 2016;95:18–31. https://doi.org/10.1016/j.cageo.2016.06.015.

[117] Zhang P, Li J, Yang X, Zhu H. Semi-automatic extraction of rock discontinuities from point clouds using the ISODATA clustering algorithm and deviation from mean elevation. Int J Rock Mech Min Sci 2018;110:76–87. https://doi.org/10.1016/j.ijrmms.2018.07.009.

[118] Gao F, Chen D, Zhou K, Niu W, Liu H. A Fast Clustering Method for Identifying Rock Discontinuity Sets. KSCE J Civ Eng 2019;23:556–66. https://doi.org/10.1007/s12205-018-1244-7.

[119] Singh SK, Banerjee BP, Lato MJ, Sammut C, Raval S. Automated rock mass discontinuity set characterisation using amplitude and phase decomposition of point cloud data. Int J Rock Mech Min Sci 2022;152:105072. https://doi.org/https://doi.org/10.1016/j.ijrmms.2022.105072.

[120] Drews T, Miernik G, Anders K, Höfle B, Profe J, Emmerich A, Bechstadt T. Validation of fracture data recognition in rock masses by automated plane detection in 3D point clouds. Int J Rock Mech Min Sci 2018;109:19–31. https://doi.org/10.1016/j.ijrmms.2018.06.023.

[121] Ge YF, Tang HM, Xia D, Wang LQ, Zhao BB, Teaway JW, Chen HZ, Zhou T.

Automated measurements of discontinuity geometric properties from a 3D-point cloud based on a modified region growing algorithm. Eng Geol 2018;242:44–54. https://doi.org/10.1016/j.enggeo.2018.05.007.

[122] Chen J, Huang H, Cohn AG, Zhang D, Zhou M. Machine learning-based classification of rock discontinuity trace: SMOTE oversampling integrated with GBT ensemble learning. Int J Min Sci Technol 2022;32:309–22. https://doi.org/https://doi.org/10.1016/j.ijmst.2021.08.004.

[123] Battulwar R, Emami E, Naghadehi MZ, Sattarvand J. Automatic Extraction of Joint Orientations in Rock Mass Using PointNet and DBSCAN. In: Proceedings of the International Symposium on Visual Computing: Advances in Visual Computing. Cham: Springer International Publishing; 2020, p. 718–27.

[124] Ben-Shabat Y, Lindenbaum M, Fischer A. Nesti-net: Normal estimation for unstructured 3D point clouds using convolutional neural networks. In: Proceedings of the IEEE/CVF Conference on Computer Vision and Pattern Recognition. Long Beach: Institute of Electrical and Electronics Engineers; 2019.p.10104–12. https://doi.org/10.1109/CVPR.2019.01035.

[125] Battulwar R, Zare-Naghadehi M, Emami E, Sattarvand J. A state-of-the-art review of automated extraction of rock mass discontinuity characteristics using three-dimensional surface models. J Rock Mech Geotech Eng 2021;13:920–36. https://doi.org/10.1016/j.jrmge.2021.01.008.

[126] Vlachopoulos N, Vazaios I, Forbes B, Carrapatoso C. Rock Mass Structural Characterization Through DFN–LiDAR–DOS Methodology. Geotech Geol Eng 2020;38:6231–44. https://doi.org/10.1007/s10706-020-01431-1.

[127] Vazaios I, Vlachopoulos N, Diederichs MS. Integration of Lidar-Based Structural Input and Discrete Fracture Network Generation for Underground Applications. Geotech Geol Eng 2017;35:2227–51. https://doi.org/10.1007/s10706-017-0240-x.

[128] Jing H, Liu X, Shao A, Wang L. Comparison and analysis of different methods for structural planes measuring in underground roadways. Energy Sources, Part A Recover Util Environ Eff 2019;00:1–15. https://doi.org/10.1080/15567036.2019.1679287.

[129] Li X, Chen J, Zhu H. A new method for automated discontinuity trace mapping on rock mass 3D surface model. Comput Geosci 2016;89:118–31. https://doi.org/10.1016/j.cageo.2015.12.010.

[130] Fekete S, Diederichs M. Integration of three-dimensional laser scanning with discontinuum modelling for stability analysis of tunnels in blocky rockmasses. Int J Rock Mech Min Sci 2013;57:11–23. https://doi.org/10.1016/j.ijrmms.2012.08.003.

[131] Baylis CNC, Kewe DR, Jones EW. Mobile drone LiDAR structural data collection and analysis. In: Proceedings of the Second International Conference on Underground Mining Technology. Perth: Australian Centre for Geomechanics; 2020, p. 325–34. https://doi.org/10.36487/ACG_repo/2035_16

[132] Turner RM, MacLaughlin MM, Iverson SR. Identifying and mapping potentially adverse discontinuities in underground excavations using thermal and multispectral UAV imagery. Eng Geol 2020;266:105470. https://doi.org/10.1016/j.enggeo.2019.105470.

[133] Singh SK, Raval S, Banerjee B. A robust approach to identify roof bolts in 3D point cloud data captured from a mobile laser scanner. Int J Min Sci Technol 2021;31:303–12. https://doi.org/10.1016/j.ijmst.2021.01.001.

[134] Lato M, Kemeny J, Harrap RM, Bevan G. Rock bench: Establishing a common repository and standards for assessing rockmass characteristics using LiDAR and photogrammetry. Comput Geosci 2013;50:106–14. https://doi.org/10.1016/j.cageo.2012.06.014.

[135] Hoek E. Putting numbers to geology—an engineer's viewpoint. Q J Eng Geol Hydrogeol 1999;32:1–19. https://doi.org/10.1144/GSL.QJEG.1999.032.P1.01.

[136] Elmo D, Yang B, Stead D, Rogers S. A Discrete Fracture Network Approach to Rock Mass Classification. vol. 125. Springer International Publishing; 2021. https://doi.org/10.1007/978-3-030-64514-4_92.

[137] Xingdong Z, Lei D, Shujing Z. Stability analysis of underground water-sealed oil storage caverns in China : A case study. Energy Explor Exploit 2020;38:2252–76. https://doi.org/10.1177/0144598720922307.

[138] Bao H, Zhang G, Lan H, Yan C, Xu J, Xu W. Geometrical heterogeneity of the joint roughness coefficient revealed by 3D laser scanning. Eng Geol 2020;265:105415. https://doi.org/10.1016/j.enggeo.2019.105415.

[139] Stille H, Palmström A. Classification as a tool in rock engineering. Tunn Undergr Sp Technol 2003;18:331–45. https://doi.org/https://doi.org/10.1016/S0886-7798(02)00106-2.

[140] Riquelme AJ, Tomás R, Abellán A. Characterization of rock slopes through slope mass rating using 3D point clouds. Int J Rock Mech Min Sci 2016;84:165–76. https://doi.org/10.1016/j.ijrmms.2015.12.008.

[141] Hoek E, Brown ET. Practical estimates of rock mass strength. Int J Rock Mech Min Sci 1997;34:1165–86. https://doi.org/10.1016/S1365-1609(97)80069-X.

[142] Hoek E, Carter TG, Diederichs MS. Quantification of the Geological Strength Index Chart. In: Proceedings of the 47th US Rock Mechanics/Geomechanics Symposium. San Francisco: American Rock Mechanics Association; 2013.p.13-672.

[143] Cai M, Kaiser PK, Uno H, Tasaka Y, Minami M. Estimation of rock mass deformation modulus and strength of jointed hard rock masses using the GSI system. Int J Rock Mech Min Sci 2004;41:3–19. https://doi.org/10.1016/S1365-1609(03)00025-X.

[144] Elmo D, Rogers S, Stead D, Eberhardt E. Discrete fracture network approach to characterise rock mass fragmentation and implications for geomechanical upscaling. Trans Institutions Min Metall Sect A Min Technol 2014;123:149–61. https://doi.org/10.1179/1743286314Y.0000000064.

[145] Monsalve JJ, Baggett J, Bishop R, Ripepi N. Application of laser scanning for rock mass characterization and discrete fracture network generation in an underground limestone mine. Int J Min Sci Technol 2019;29:131–7. https://doi.org/10.1016/j.ijmst.2018.11.009.

[146] Mah J, Samson C, McKinnon SD, Thibodeau D. 3D laser imaging for surface roughness

analysis. Int J Rock Mech Min Sci 2013;58:111–7. https://doi.org/10.1016/j.ijrmms.2012.08.001.

[147] Lindenbergh R, Pietrzyk P. Change detection and deformation analysis using static and mobile laser scanning. Appl Geomatics 2015;7:65–74. https://doi.org/10.1007/s12518-014-0151-y.

[148] Slaker B, Westman E. Identifying underground coal mine displacement through field and laboratory laser scanning. J Appl Remote Sens 2014;8:083544. https://doi.org/10.1117/1.jrs.8.083544.

[149] Navarro J, Segarra P, Sanchidrián JA, Castedo R, López LM. Assessment of drilling deviations in underground operations. Tunn Undergr Sp Technol 2019;83:254–61. https://doi.org/10.1016/j.tust.2018.10.003.

[150] Walton G, Diederichs MS, Weinhardt K, Delaloye D, Lato MJ, Punkkinen A. Change detection in drill and blast tunnels from point cloud data. Int J Rock Mech Min Sci 2018;105:172–81. https://doi.org/10.1016/j.ijrmms.2018.03.004.

[151] Evanek N, Slaker B, Iannacchione A, Miller T. LiDAR mapping of ground damage in a heading re-orientation case study. Int J Min Sci Technol 2021;31:67–74. https://doi.org/10.1016/j.ijmst.2020.12.018.

[152] Iannacchione A, Miller T, Esterhuizen G, Slaker B, Murphy M, Cope N, Thayer S. Evaluation of stress-control layout at the Subtropolis Mine, Petersburg, Ohio. Int J Min Sci Technol 2020;30:77–83. https://doi.org/10.1016/j.ijmst.2019.12.009.

[153] Slaker B, Murphy M, Rashed G, Gangrade V, Van Dyke M, Minoski T, Floyd K. Monitoring of Multiple-level Stress Interaction at Two Underground Limestone Mines. Mining, Metall Explor 2021;38:623–33. https://doi.org/10.1007/s42461-020-00345-z.

[154] Kukutsch R, Kajzar V, Waclawik P, Nemcik J. Application of the Terrestrial 3D Laser Scanning in Room and Pillar Trial at CSM Mine. In: Proceedings of the ISRM International Symposium-10th Asian Rock Mechanics Symposium. Singapore: International Society for Rock Mechanics and Rock Engineering; 2018.p.1–9.

[155] Gálai B, Benedek C. Change Detection in Urban Streets by a Real Time Lidar Scanner and MLS Reference Data. In: Proceedings of the International Conference Image Analysis and Recognition. Cham: Springer International Publishing; 2017.p. 210–20.

[156] Kromer RA, Abellán A, Hutchinson DJ, Lato M, Chanut MA, Dubois L, Jaboyedoff M. Automated terrestrial laser scanning with near-real-time change detection - Monitoring of the Séchilienne landslide. Earth Surf Dyn 2017;5:293–310. https://doi.org/10.5194/esurf-5-293-2017.

[157] Dawn T. Technologies of ground support monitoring in block caving operations. In: Proceedings of the Ninth International Symposium on Ground Support in Mining and Underground Construction. Sudbury: Australian Centre for Geomechanics; 2019.p.109–22.

[158] MSHA. Mineral Resources-Mine Safety and Health Administration. North Capitol St NW: U.S. Government Publishing Office; 2019. https://www.govinfo.gov/content/pkg/CFR-2019-title30-vol1/pdf/CFR-2019-title30-vol1.pdf

[159] Pinpin L, Wenge Q, Yunjian C, Feng L. Application of 3D Laser Scanning in Underground Station Cavity Clusters. Adv Civ Eng 2021;2021. https://doi.org/10.1155/2021/8896363.

[160] Puente I, Akinci B, González-Jorge H, Díaz-Vilariño L, Arias P. A semi-automated method for extracting vertical clearance and cross sections in tunnels using mobile LiDAR data. Tunn Undergr Sp Technol 2016;59:48–54. https://doi.org/10.1016/j.tust.2016.06.010.

[161] MQSH. Mining and Quarrying Safety and Health Regulation 2017 2017.

[162] Benton DJ, Chambers AJ, Raffaldi MJ, Finley SA, Powers MJ. Close-range photogrammetry in underground mining ground control 2016;997707:997707. https://doi.org/10.1117/12.2236691.

[163] Gallwey J, Eyre M, Coggan J. A machine learning approach for the detection of supporting rock bolts from laser scan data in an underground mine. Tunn Undergr Sp Technol 2021;107:103656. https://doi.org/10.1016/j.tust.2020.103656.

[164] Singh SK, Raval S, Banerjee B. Roof bolt identification in underground coal mines from 3D point cloud data using local point descriptors and artificial neural network. Int J Remote Sens 2021;42:367–77. https://doi.org/10.1080/2150704X.2020.1809734.

[165] Saydam S, Liu B, Li B, Zhang W, Singh SK, Raval S. Effective rock bolt detection in underground tunnels. IEEE Access 2021;PP:1–1. https://doi.org/10.1109/access.2021.3120207.

[166] Bjureland W, Johansson F, Spross J, Larsson S. Influence of spatially varying thickness on load-bearing capacity of shotcrete. Tunn Undergr Sp Technol 2020;98:103336. https://doi.org/10.1016/j.tust.2020.103336.

[167] Fekete S, Diederichs M, Lato M. Geotechnical and operational applications for 3-dimensional laser scanning in drill and blast tunnels. Tunn Undergr Sp Technol 2010;25:614–28. https://doi.org/10.1016/j.tust.2010.04.008.

[168] Lato MJ, Diederichs MS. Mapping shotcrete thickness using LiDAR and photogrammetry data: Correcting for over-calculation due to rockmass convergence. Tunn Undergr Sp Technol 2014;41:234–40. https://doi.org/10.1016/j.tust.2013.12.013.

[169] Martínez-Sánchez J, Puente I, González-Jorge H, Riveiro B, Arias P. Automatic thickness and volume estimation of sprayed concrete on anchored retaining walls from terrestrial LIDAR data. Int Arch Photogramm Remote Sens Spat Inf Sci - ISPRS Arch 2016;41:521–6. https://doi.org/10.5194/isprsarchives-XLI-B5-521-2016.

[170] Wrock MR, Nokleby SB. Robotic shotcrete thickness estimation using fiducial registration. In: Proceedings of the ASME 2018 International Design Engineering Technical Conferences and Computers and Information in Engineering Conference. Quebec City: The American Society of Mechanical Engineers; 2018.p.1–10. https://doi.org/10.1115/DETC201885688.

[171] Abdellah WRE, Hefni MA, Ahmed HM. Factors Influencing Stope Hanging Wall Stability and Ore Dilution in Narrow-Vein Deposits: Part 1. Geotech Geol Eng 2020;38:1451–70. https://doi.org/10.1007/s10706-019-01102-w.

[172] Clark L, Pakalnis R. An empirical design approach for estimating unplanned dilution

from open stope hangingwalls and footwalls. Presentation at 99th Canadian Institute of Mining annual conference. Vancouver: Canadian Institute of Mining, Metallurgy and Petroleum; 1997.

[173] Janiszewski M, Pontow S, Rinne M. Industry Survey on the Current State of Stope Design Methods in the Underground Mining Sector. Energies 2022;15. https://doi.org/10.3390/en15010240.

[174] Henning JG, Mitri HS. Assessment and control of ore dilution in long hole mining: Case studies. Geotech Geol Eng 2008;26:349–66. https://doi.org/10.1007/s10706-008-9172-9.

[175] Amedjoe CG, Agyeman J. Assessment of effective factors in performance of an open stope using cavity monitoring system data: A case study. J Geol Min Res 2015;7:19–30. https://doi.org/10.5897/jgmr2014.0215.

[176] Lee SJ, Choi SO. Analyzing the Stability of Underground Mines Using 3D Point Cloud Data and Discontinuum Numerical Analysis. Sustain 2019;11. https://doi.org/10.3390/su11040945.

[177] Oparin VN, Yushkin VF, Klimko VK, Rublev DE, Izotov AS, Ivanov A V. Analytical Description of Surface of Blasting-Formed Underground Cavities by Laser Scanning Data. J Min Sci 2018;53:789–800. https://doi.org/10.1134/S1062739117042789.

[178] Aubertin JD, Hutchinson DJ, Diederichs M. Horizontal single hole blast testing – Part 1 : Systematic measurements using TLS surveys. Tunn Undergr Sp Technol 2021;114:103985. https://doi.org/10.1016/j.tust.2021.103985.

[179] Campbell AD, Thurley MJ. Application of laser scanning to measure fragmentation in underground mines. Min Technol 2017;126:1–8. https://doi.org/10.1080/14749009.2017.1296668.

[180] Yuan Z, Ban X, Han F, Zhang X, Yin S, Wang Y. Integrated three-dimensional visualization and soft-sensing system for underground paste backfilling. Tunn Undergr Sp Technol 2022;127:104578. https://doi.org/10.1016/j.tust.2022.104578.

[181] Watson C, Marshall J. Estimating underground mine ventilation friction factors from low density 3D data acquired by a moving LiDAR. Int J Min Sci Technol 2018;28:657–62. https://doi.org/10.1016/j.ijmst.2018.03.009.

[182] Kot T, Novak P, Babjak J. System for creation and display of 3D maps of coal mines. In: Proceedings of the 17th Coal Operators' Conference. Wollongong: University of Wollongong; 2017.p.117–24.

[183] Johnson CP. A Guide to Surface Features Related to Underground Coal Mining. Master's dissertation. Seattle: University of Washington; 2013.p.39.